\documentclass[conference]{IEEEtran}
\IEEEoverridecommandlockouts
\usepackage{cite}
\usepackage{amsmath,amssymb,amsfonts}
\usepackage{algorithmic}
\usepackage{graphicx}
\usepackage{textcomp}
\usepackage{xcolor}
\usepackage{setspace}
\usepackage{balance}
\usepackage{makecell}
\usepackage{amsthm}
\usepackage{graphicx}
\usepackage{subfigure}
\usepackage{color}
\usepackage{ifpdf}
\usepackage{CJK}
\usepackage{epstopdf}
\usepackage{amsfonts}
\usepackage{xspace}
\usepackage{algorithm}
\usepackage{url}
\usepackage{bbding}
\usepackage{booktabs}
\usepackage{tikz}
\newcommand*{\circled}[1]{\lower.8ex\hbox{\tikz\draw (0pt, 0pt)%
    circle (.47em) node {\makebox[0.4em][c]{\small #1}};}}

\def\ie{\textit{i.e.}\xspace}
\def\etal{\textit{et al.}\xspace}

\def\eg{\textit{e.g.}\xspace}

\def\wrt{\textit{w.r.t.}\xspace}

\def\method{MergeSFL\xspace}
\def\fmbr{strategies\xspace}

\begin{document}

\title{MergeSFL: Split Federated Learning with Feature Merging and Batch Size Regulation}

\author{
\IEEEauthorblockN{
    Yunming Liao$^{1,2}$~~$^{\ast}$Yang Xu$^{1,2}$~~Hongli Xu$^{1,2}$~~Lun Wang$^{1,2}$~~Zhiwei Yao$^{1,2}$~~Chunming Qiao$^{3}$
}
\IEEEauthorblockA{
    $^{1}$School of Computer Science and Technology, University of Science and Technology of China\\
    $^{2}$Suzhou Institute for Advanced Research, University of Science and Technology of China\\
    $^{3}$Department of Computer Science and Engineering, University at Buffalo, the State University of New York\\
    \{ymliao98, wanglun0, zhiweiyao,\}@mail.ustc.edu.cn,~\{xuyangcs, xuhongli\}@ustc.edu.cn,~qiao@buffalo.edu
}
\IEEEcompsocitemizethanks{
    \IEEEcompsocthanksitem *Corresponding author.
}
}

\maketitle

\thispagestyle{plain} 

\begin{abstract}
Recently, federated learning (FL) has emerged as a popular technique for edge AI to mine valuable knowledge in edge computing (EC) systems.
To boost the performance of AI applications, large-scale models have received increasing attention due to their excellent generalized abilities. 
However, training and transmitting large-scale models 
will incur significant computing and communication burden on the resource-constrained workers, and the exchange of entire models may violate model privacy.
To relax the burden of workers and protect model privacy, split federated learning (SFL) has been released by integrating both data and model parallelism.
Despite resource limitations, SFL also faces two other critical challenges in EC systems, \ie, statistical heterogeneity and system heterogeneity.
In order to address these challenges, we propose a novel SFL framework, termed \textit{\method}, by incorporating feature merging and batch size regulation in SFL.
Concretely, feature merging aims to merge the features from workers into a mixed feature sequence, which is approximately equivalent to the features derived from IID data and is employed to promote model accuracy. 
While batch size regulation aims to assign diverse and suitable batch sizes for heterogeneous workers to improve training efficiency.
Moreover, \method explores to jointly optimize these two \fmbr upon their coupled relationship to better enhance the performance of SFL.
Extensive experiments are conducted on a physical platform with 80 NVIDIA Jetson edge devices, and the experimental results show that \method can improve the final model accuracy by 5.82\% to 26.22\%, with a speedup by about 1.39$\times$ to 4.14$\times$, compared to the baselines.

\end{abstract}


\section{Introduction}\label{sec:intro}

As an emerging and popular technique in edge AI, federated learning (FL) is proposed to train a globally-shared model through collaboration among workers (\eg, IoT devices) in the data-parallel fashion \cite{hard2018federated, kairouz2019advances, han2021legodnn, mcmahan2017communication, chen2023enhancing, wang2023distribution}.
Under coordination of the parameter server (PS), participating workers periodically train deep learning (DL) models on their local datasets, and then push the models to the PS for global aggregation without exposing their raw data.
FL has been leveraged by Google to develop the Gboard application with improved user experience in a privacy-preserving manner \cite{yang2018applied}.
To boost the performance of AI applications or services, it is usually practical and effective to augment the parameters of DL models \cite{nakkiran2021deep, zhang2021elf}.
However, training large-scale models is challenging for resource-constrained workers due to their hardware limitations of CPU and memory \cite{vepakomma2018split, park2021communication, pal2021server}. 
Additionally, transmitting large-scale models between workers and the PS incurs significant communication latency, and the exchange of entire models may violate model privacy \cite{thapa2022splitfed, han2021accelerating}.

To mitigate the computing/communication burden on the resource-constrained workers and better protect model privacy, split federated learning (SFL) has been proposed by incorporating both data parallelism and model parallelism \cite{pal2021server, thapa2022splitfed, han2021accelerating, abedi2020fedsl}.
SFL splits an entire model into two submodels, \ie, bottom model and top model, at a certain neural layer, termed the split layer.
The bottom model (close to the input) is trained on the workers, while the training of the top model (close to the output) is offloaded to the relatively resource-rich PS.
Thus, SFL significantly reduces the computing load on the workers, which makes it feasible and efficient to train larger-scale models \cite{han2021accelerating, oh2022locfedmix, liao2023accelerating}.
Different from typical FL, only the bottom models plus the features (also called activations or smashed data) and the gradients of the split layer are exchanged between workers and the PS.
Since the size of the bottom model or features/gradients is much smaller than that of an entire model, the communication load is greatly reduced.
For example, the size of a 16-layer VGG16 \cite{simonyan2014very} is about 321MB, whereas the sizes of its bottom model and the features/gradients (with batch size of 64) separately are about 56MB and 3MB, when splitting the model at the 13th layer.
As workers only have access to the bottom models and process their data locally, the privacy of user data and models is effectively protected \cite{thapa2022splitfed, abedi2020fedsl}.
Besides, the existing privacy preserving techniques such as Differential Privacy \cite{thapa2021advancements, wu2023federated, ghazi2021deep} and Homomorphic Encryption \cite{yang2023dynamic} can be employed to further protect privacy of features/gradients in SFL.


Although SFL provides the aforementioned advantages, it still suffers from two other critical challenges in practical applications.
1) \textbf{\textit{Statistical Heterogeneity.}} The workers always collect local data based on their locations and/or user preferences \cite{wang2022accelerating, xie2023federatedscope, liu2022enhancing, gui2023sk, liao2023decentralized}.
Besides, the raw data of workers is not shared with others to prevent privacy leakage, resulting in non-independent and identically distributed (non-IID) data across all workers \ie, statistical heterogeneity \cite{zhao2022fedgan, liao2023adaptive, zhuang2022divergence, shin2022fedbalancer, wang2022enhancing}.
The non-IID data decelerates the convergence rate and even compromises the accuracy of the trained models \cite{zhao2018federated, li2022pyramidfl}.
2) \textbf{\textit{System Heterogeneity.}} In EC systems, workers commonly possess varying and limited capabilities \cite{zhang2018adaptive, li2022pyramidfl}.
The computing capabilities (\eg, CPU frequency) and communication capabilities (\eg, bandwidth, throughput) of workers could differ from each other by more than tenfold times \cite{chen2022decentralized, lai2021oort, xu2022adaptive}.
System heterogeneity poses significant influences on synchronous training processes, as fast workers may be forced to wait for slow ones, leading to increased waiting time and decreased training efficiency.

So far, the existing SFL works have mainly focused on training a large-scale DL model on resource-constrained workers, without simultaneously resolving the aforementioned system and statistical heterogeneity \cite{thapa2022splitfed, han2021accelerating, pal2021server}.
For instance, SplitFed \cite{thapa2022splitfed} is the first to demonstrate the feasibility of SFL, and aggregates bottom models after each local updating.
Such frequent aggregation results in high network traffic consumption.
To save the traffic consumption, LocFedMix-SL \cite{oh2022locfedmix} proposes to reduce the aggregation frequency of bottom models, but it cannot fully utilize the capacities of heterogeneous workers.
As an advanced solution, AdaSFL \cite{liao2023accelerating} assigns adaptive and diverse batch sizes for different workers to address system heterogeneity, but still cannot deal with the statistical heterogeneity.
Prior to the emergence of SFL, many solutions to address the heterogeneity challenges \cite{li2022data,xie2023federatedscope,luo2022tackling,arisdakessian2023towards,li2022pyramidfl,lai2021oort, liu2019accelerate, tyagi2020taming, ma2021adaptive} have been studied in typical FL scenarios.
In order to alleviate the negative effect of system heterogeneity, some works \cite{tyagi2020taming, ma2021adaptive, liu2019accelerate} investigate to optimize the batch sizes of different workers. 
In addition, other works \cite{luo2022tackling,li2022pyramidfl,lai2021oort} propose to employ worker selection to simultaneously address heterogeneity issues. 
For example, PyramidFL \cite{li2022pyramidfl} proposes a fine-grained worker selection strategy that focuses on the divergence between the selected and remaining workers to fully exploit the computing resource and data of different workers.
However, those FL researches can not be directly applied for SFL, since workers maintaining only the bottom models in SFL must complete the whole training procedure by interacting with the top model residing on the PS.

To expand the ability of addressing the heterogeneity challenges for SFL, we review the distinct properties of SFL compared to those of FL, and propose a novel SFL framework, termed \textit{\method}.
The design of \method is based on two fundamental observations.
1) In SFL, the top model can be regarded as a classifier \cite{chen2022decentralized, zhang2022splitavg}, 
and the features derived from non-IID data always mislead the convergence direction of the top model, leading to the degradation of model accuracy.
As illustrated in Section \ref{Sec:feature_merge}, if we merge the features from different workers to form a mixed feature sequence, which is approximately equivalent to the features derived from an IID mini-batch, the top model will be updated along the reasonably optimal direction.
2) Inspired by previous FL works \cite{ma2021adaptive, tyagi2020taming}, assigning appropriate batch sizes for different workers will help to accommodate to their diverse capacities.
For example, the workers with high computing capacities are assigned with large batch sizes \cite{ma2021adaptive}, thus the time consumption of performing forward and backward propagation across workers can be essentially the same, and the system heterogeneity is expected to be addressed.

Motivated by the above insights, \method explores to build up an efficient SFL system by combining feature merging and batch size regulation.
The difficulty of system design lies in \textit{the interactions between feature merging and batch size regulation}.
On one hand, to make full use of local data across workers, it is desirable to collect enough features (indicating large batch sizes) from different workers at each iteration.
However, considering resource limitation and system heterogeneity, \method needs to assign appropriate (but relatively small) batch sizes for the workers to balance their training time.
On the other hand, since the merged feature sequence is composed of the mini-batches from different workers, given the workers with diverse batch sizes, \method should dynamically select suitable workers and arrange their features to form a large IID mini-batch.
\textit{Only by jointly optimizing feature merging and batch size regulation, does \method contribute to well tackling the heterogeneity challenges, and realize efficient SFL}, which, to our best knowledge, has not been investigated in existing literature.
In a nutshell, the main contributions of this paper are summarized as follows:
\begin{itemize}
    \item We review the characteristic properties of SFL, and propose a novel SFL framework, termed \method, which incorporates feature merging and batch size regulation to overcome the challenges of system and statistical heterogeneity.
    \item We analyze joint influence of feature merging and batch size regulation on training performance and obtain their coupled relationship.
    Then, \method dynamically selects and arranges a part of workers under the heterogeneity restrictions,
    thereby promoting model accuracy as well as training efficiency.
    \item The performance of \method is evaluated through a physical platform with totally 80 NVIDIA Jetson edge devices.
    The experimental results show that \method improves the final model accuracy by 5.82\% to 26.22\%, with a speedup by about 1.39$\times$ to 4.14$\times$, compared to the baselines.
\end{itemize}

The rest of the paper is organized as follows.
Section \ref{sec:prelim} presents the background of split federated learning, and introduces the motivations of our \method system.
Section \ref{sec:overview} illustrates the overview of \method.
Then we elaborate the detailed design of 
\method in Section \ref{sec:design}.
The experimental evaluation is presented in Section \ref{sec:evaluation}.
We review some related works in Section \ref{sec:relwork} and conclude the paper in Section \ref{sec:conclusion}.

\section{Background and Motivation}\label{sec:prelim}
\begin{figure*}[t]
\centering
\includegraphics[width=0.8\linewidth]{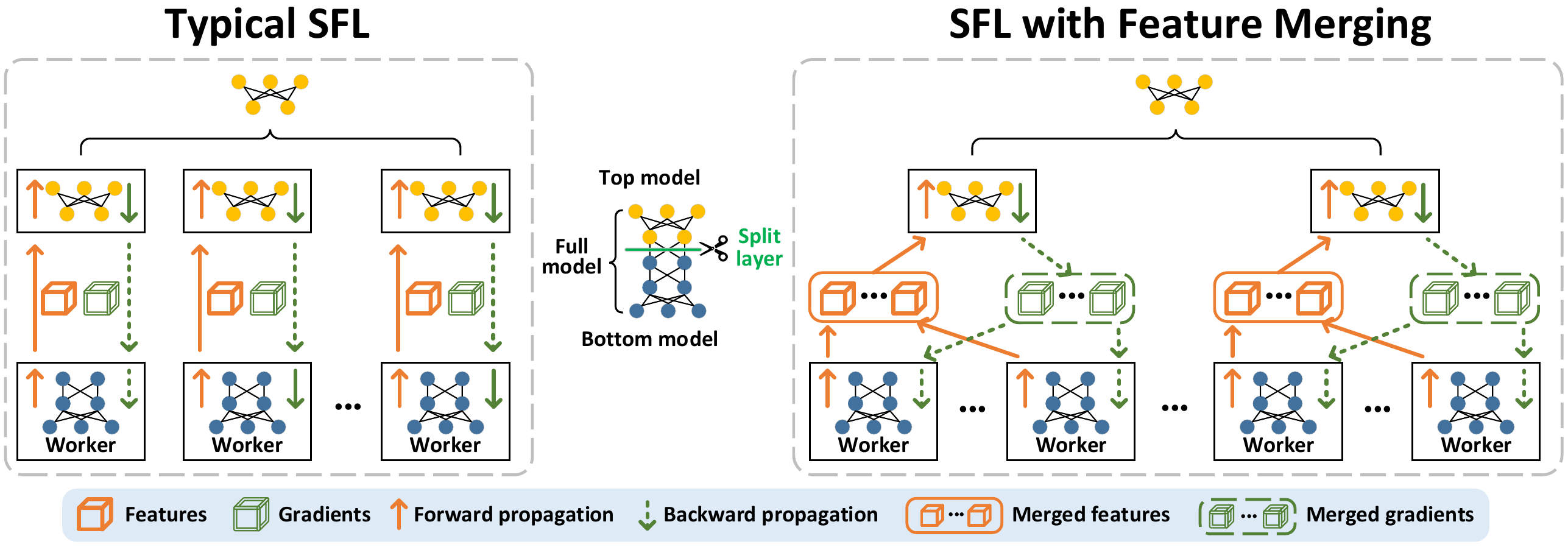}
\vspace{-0.2cm}
\caption{Illustration of typical SFL (left) and SFL with feature merging (right). }
\label{fig:SFL}
\vspace{-0.4cm}
\end{figure*}

\subsection{Split Federated Learning}
Considering an EC system with a parameter server (PS) and totally $N$ workers, split federated learning (SFL) is proposed to perform deep learning tasks through a loose federation of workers coordinated by the PS.
The basic idea of SFL is to split the model $\boldsymbol{w}$ into two submodels at the split layer, \ie, $\boldsymbol{w}=[\boldsymbol{w}_b, \boldsymbol{w}_p]$, where $\boldsymbol{w}_b$ denotes the bottom (sub-)model and $\boldsymbol{w}_p$ is the top (sub-)model.
For ease of description, we take the CNN model as an example. 
The bottom model usually consists of the input layer and convolutional layers whereas the top model includes fully-connected layers and the output layer.
In SFL, the PS maintains the top model $\boldsymbol{w}_p$, while each worker $i$ ($\in [N]$) trains a bottom model $\boldsymbol{w}_{b,i}$ using its local data $\mathbb{D}_i$.
Besides, the workers complete the whole training procedure by interacting the top model residing on the PS.
The goal of SFL is to find the optimal model $\boldsymbol{w}^* = [\boldsymbol{w}^*_b, \boldsymbol{w}^*_p]$ that minimizes the loss function $F(\boldsymbol{w})$ as follows:
\vspace{-0.4cm}
\begin{equation}\label{equ:opti}
    \min _{\boldsymbol{w}=[\boldsymbol{w}_b, \boldsymbol{w}_p]} F(\boldsymbol{w}) \triangleq F_p(\boldsymbol{w}_p) + \frac{1}{N} \sum_{i=1}^{N} F_{b,i}(\boldsymbol{w}_{b,i})
    \vspace{-0.1cm}
\end{equation}
where $F_{b,i}(\boldsymbol{w}_{b,i})$ and $F_p(\boldsymbol{w}_p)$ separately denote the loss functions of bottom model $\boldsymbol{w}_{b,i}$ on worker $i$ and top model $\boldsymbol{w}_p$.
Due to the intrinsic complexity of most deep learning tasks, it is usually challenging to obtain a closed-form solution of Eq. \eqref{equ:opti}.
Nevertheless, Eq. \eqref{equ:opti} can be solved by the mini-batch stochastic gradient descent (SGD) algorithms in SFL \cite{thapa2022splitfed, han2021accelerating}.
For ease of expression, some important notations in this paper are listed in Table \ref{tab:notation}.

\begin{table}[t]
\caption{Key Notations.}
\centering
\begin{spacing}{0.8}
    \begin{tabular}{cl}
        \toprule
        \textbf{Notation} & \makecell[c]{\textbf{Semantics}} \\
        \midrule
        $N$ & number of workers\\
        \specialrule{0em}{1pt}{1pt}
        $\mathbb{D}_i$  & local dataset of worker $i$ \\
        \specialrule{0em}{1pt}{1pt}
        $\boldsymbol{w}_p^h$  & top model model in round $h$\\
        \specialrule{0em}{1pt}{1pt}
        $\boldsymbol{w}_{b,i}^h$  & bottom model on worker $i$ in round $h$\\
        \specialrule{0em}{1pt}{1pt}
        $F_p(\boldsymbol{w}_p)$  & loss function of the top model\\
        \specialrule{0em}{1pt}{1pt}
        $F_{b_i}(\boldsymbol{w}_{b,i})$  & loss function of bottom model on worker $i$\\
        \specialrule{0em}{1pt}{1pt}
        $d_i^h$ & batch size of worker $i$ in round $h$\\
        \specialrule{0em}{1pt}{1pt}
        $t_i^h$ & duration time of worker $i$ in round $h$\\
        \specialrule{0em}{1pt}{1pt}
        $\mathcal{W}^h$ & average waiting time of round $h$\\
        \specialrule{0em}{1pt}{1pt}
        $B^h$ & the available ingress bandwidth of the PS \\ &in round $h$\\
        \specialrule{0em}{1pt}{1pt}
        $\mathcal{S}^h$ & the worker set for feature merging \\ &and model training in round $h$\\
        \specialrule{0em}{1pt}{1pt}
        $\mu_i^h$ & computing time of processing one data \\ &sample in round $h$ on worker $i$\\
        \specialrule{0em}{1pt}{1pt}
        $\eta_i^h$ & communication time of transmitting one \\ & data sample in round $h$ on worker $i$\\
        \specialrule{0em}{1pt}{1pt}
        $\textbf{V}_i$ & the label distribution of worker $i$\\
        \specialrule{0em}{1pt}{1pt}
        $\Phi_h$ & the label distribution of data from \\ & workers in $\mathcal{S}^h$\\
        \specialrule{0em}{1pt}{1pt}
        $K_i$ & the participating frequency of worker $i$\\
        \specialrule{0em}{1pt}{1pt}
        \bottomrule
    \end{tabular}
\end{spacing}
\label{tab:notation}
\vspace{-0.2cm}
\end{table}

The basic training process of SFL involves three main stages, \ie, forward/backward propagation of the worker-specific bottom models, forward/backward propagation of the top model, and global aggregation of bottom models on the PS.
Firstly, each worker performs forward propagation with a batch of data samples, and delivers the features (also called \textit{smashed data}) of the split layer to the PS.
Subsequently, the PS performs forward/backward propagation to update the top model.
Then, the PS sends the backpropagated gradients back to the workers for updating the bottom models by proceeding backward propagation.
Such a complete process of forward/backward propagation is regarded as a local iteration.
After several local iterations, the PS aggregates the bottom models from all workers and sends the aggregated bottom model back to the workers for further training.
The above whole training process is regarded as a communication round. 
Let $\boldsymbol{w}_{b,i}^{h,k}$ denote the bottom model of worker $i$ at the $k$-th iteration in round $h$.
Then, after one iteration, the bottom model is updated as:
\vspace{-0.1cm}
\begin{equation}
    \boldsymbol{w}_{b,i}^{h,k+1}=\boldsymbol{w}_{b,i}^{h,k}- \eta \widetilde{\nabla} F_{b,i}(\boldsymbol{w}_{b,i}^{h,k})
\end{equation}
Herein, $\widetilde{\nabla} F_{b,i}(\boldsymbol{w}_{b,i}^{h,k})=\frac{1}{|D_{i}|} \sum_{x \in D_{i}} \nabla \ell(x ; \boldsymbol{w}_{b,i}^{h,k})$ is the gradient for a certain mini-batch $D_{i}$ with size $d_i=|D_{i}|$, and $\nabla \ell(x ; \boldsymbol{w}_{b,i}^{h,k})$ denotes the stochastic gradient given the bottom model $\boldsymbol{w}_{b,i}^{h,k}$ and input data sample $x$.
Let $\boldsymbol{w}_{p}^{h,k}$ denote the top model at the $k$-th iteration in round $h$.
Accordingly, the process of updating the top model at one iteration is expressed as follows:
\vspace{-0.1cm}
\begin{equation}\label{server-update}
    \boldsymbol{w}_{p}^{h,k+1}=\boldsymbol{w}_{p}^{k} - \eta \sum\limits_{i=1}^N\! \frac{d_i}{\sum_{i=1}^N d_i} \widetilde{\nabla} F_{p,i}(\boldsymbol{w}_{p}^{h,k})
\end{equation}
where $\widetilde{\nabla} F_{p,i}(\boldsymbol{w}_{p}^{h,k})=\frac{1}{|D_{i}|} \sum_{x \in D_{i}} \nabla \ell(\boldsymbol{w}_{b,i}^{h,k}(x); \boldsymbol{w}_{p}^{h,k})$ is the gradient of top loss function, and $\nabla \ell(\boldsymbol{w}_{b,i}^{h,k}(x); \boldsymbol{w}_{p}^{h,k})$ denotes the stochastic gradient for the top model $\boldsymbol{w}_{p}^{h,k}$ and the output of the bottom model $\boldsymbol{w}_{b,i}^{h,k}$ (\ie, features), given the input data sample $x$.
After local updating, the PS receives and aggregates the bottom models from all workers:
\vspace{-0.1cm}
\begin{equation}\label{aggregation}
    \boldsymbol{w}_{b}^{h+1}=\frac{1}{N}\sum_{i=1}^N \boldsymbol{w}_{b,i}^{h+1}
    \vspace{-0.1cm}
\end{equation}

\vspace{0.1cm}
\subsection{Importance of Feature Merging} \label{Sec:feature_merge}
Different from the assumption in traditional centralized training with IID data distribution, the distributions of local data on geographically diverse workers vary significantly, which causes the non-IID issue (\ie, statistical heterogeneity) and deteriorates the training performance \cite{mcmahan2017communication, zhuang2022divergence}.
In typical SFL (\eg, LocFedMix-SL \cite{oh2022locfedmix}), the PS directly applies the features of each worker to complete the forward/backward propagation of the top model, and sends the corresponding gradients back to the workers in sequence.
However, the features of different workers with non-IID data may hinder the model from being updated along the optimal directions, which reduces model accuracy.
To tackle the non-IID issue, we introduce the strategy of \textit{feature merging} in typical SFL.
The difference between typical SFL and SFL with feature merging is illustrated in Fig. \ref{fig:SFL}.
The idea of feature merging is to merge the features (with small batch sizes) from different workers to form one mixed feature sequence (with a large batch size), which is approximately equivalent to the feature derived from IID mini-batch and is utilized to conduct forward/backward propagation of the top model.
Subsequently, the PS segments the mixed large-size gradient into multiple small-size gradients corresponding to each worker, and then dispatches the gradients back to the workers to update the bottom models.
Considering the advantage of feature merging, both the top model and bottom models will be updated along the relatively right directions in the optimization space, which contributes to improving model accuracy.

\begin{figure}[t]
\centering
\subfigure[Test Accuracy]
{
    \includegraphics[width=0.425\linewidth,height=2.8cm]{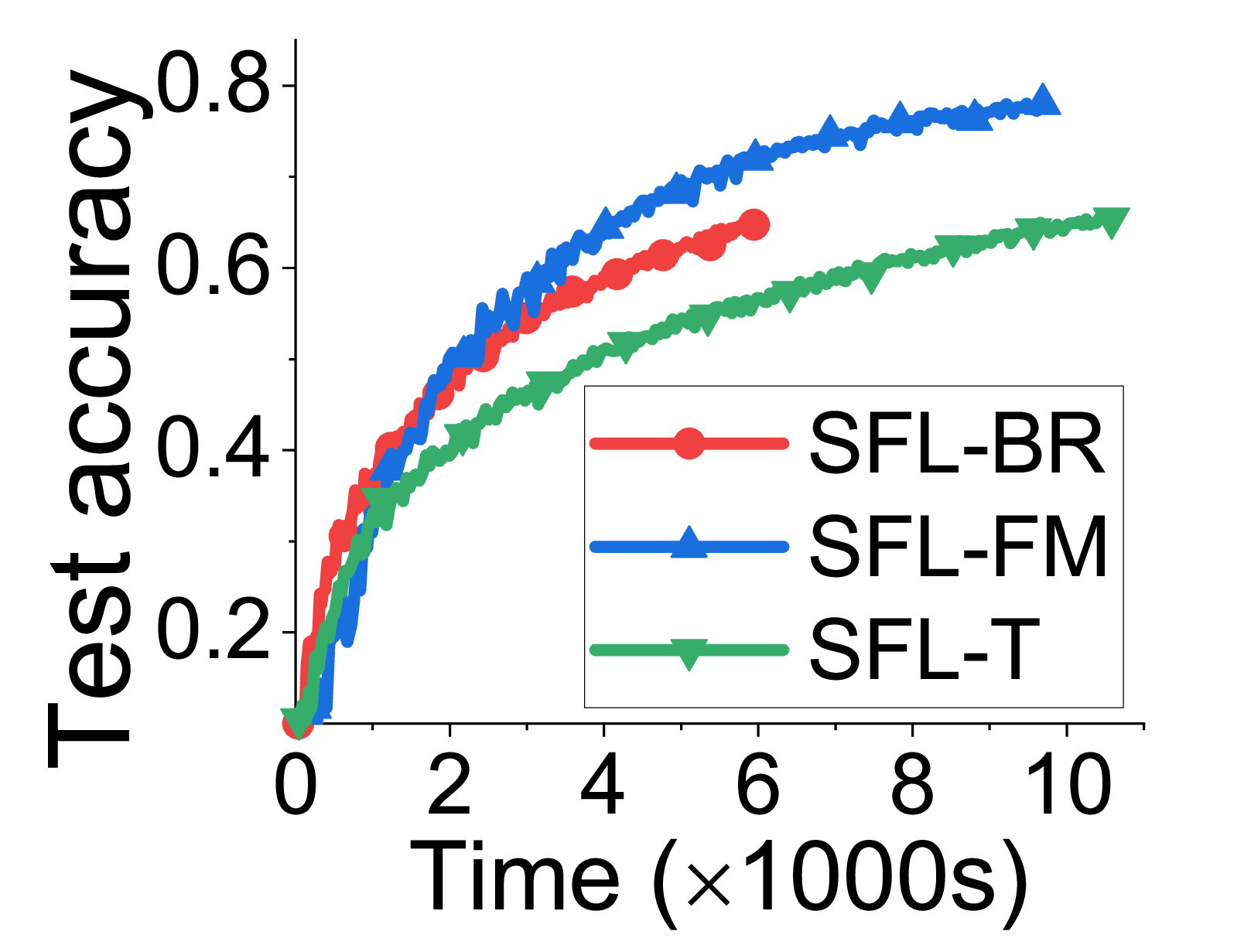}
    \label{fig:selection_time_acc}
}\quad 
\subfigure[Average Waiting time]
{
    \includegraphics[width=0.425\linewidth,height=2.8cm]{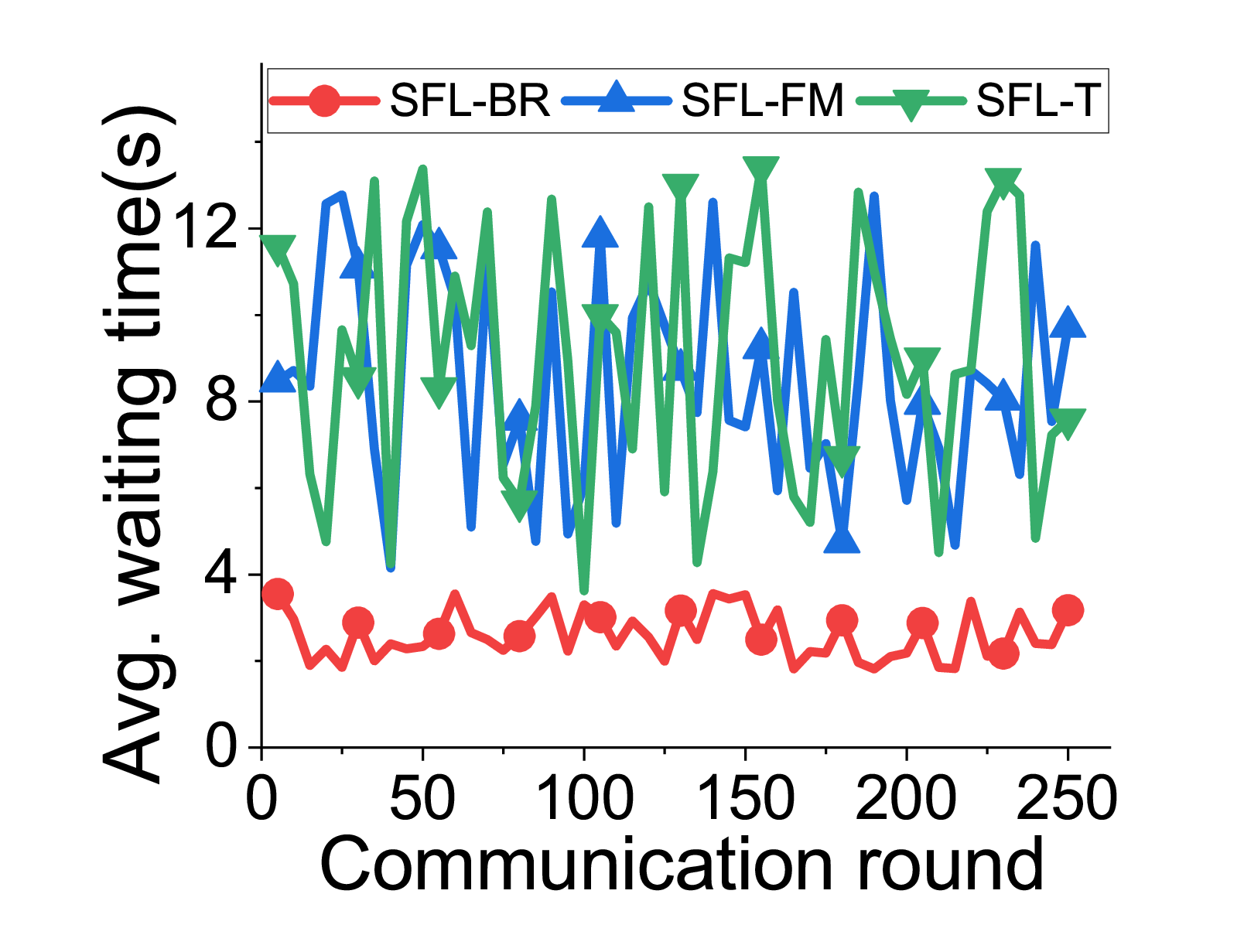}
    \label{fig:selection_waiting_time}
}
\vspace{-0.1cm}
\caption{Test accuracy and average waiting time of three approaches with non-IID data.}
\label{fig:selection_test}
\vspace{-0.5cm}
\end{figure}

\begin{figure}[t]
\centering
\subfigure[Training Time]
{
    \includegraphics[width=0.425\linewidth,height=2.8cm]{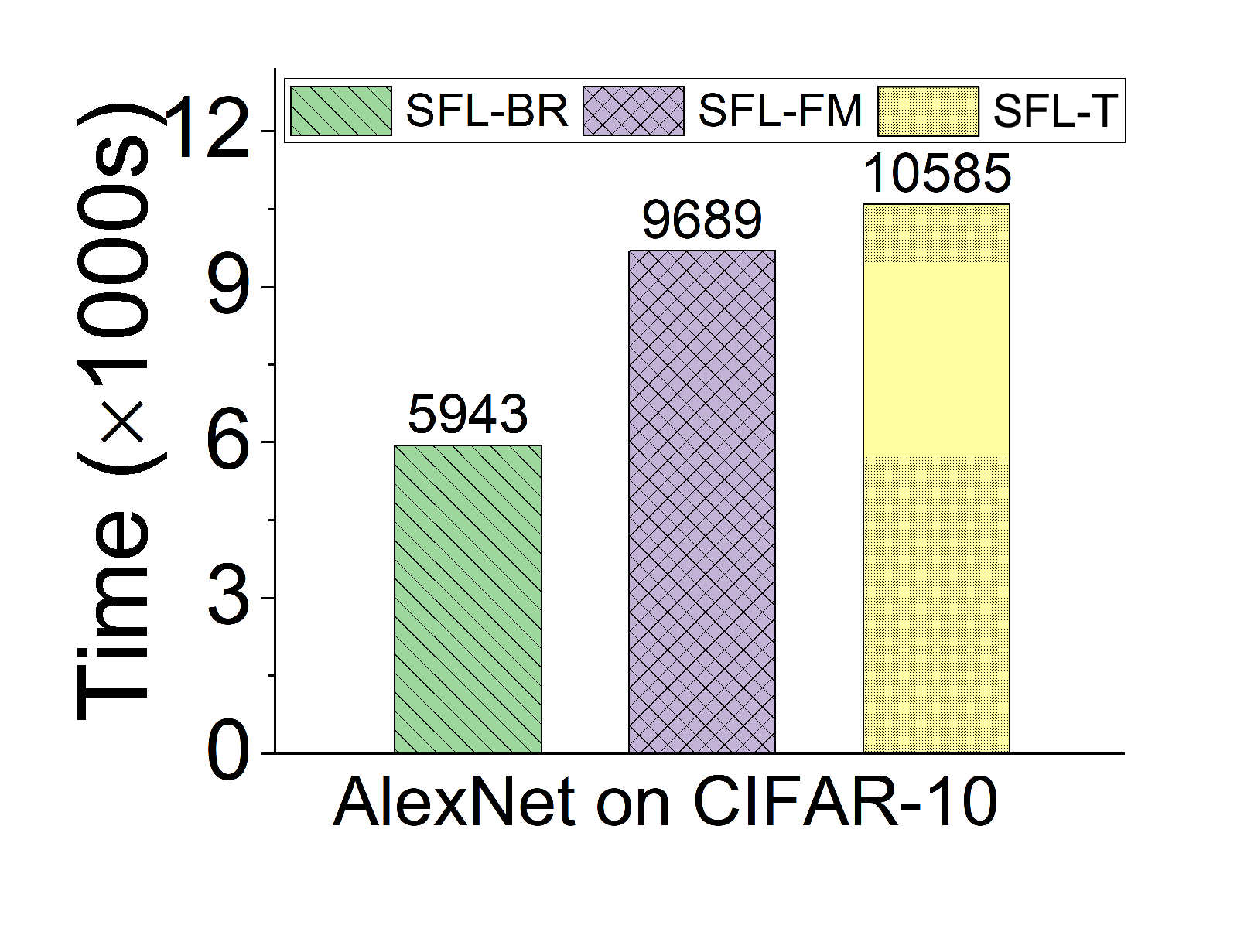}
    \label{fig:selection_time}
}\quad 
\subfigure[Test Accuracy]
{
    \includegraphics[width=0.425\linewidth,height=2.8cm]{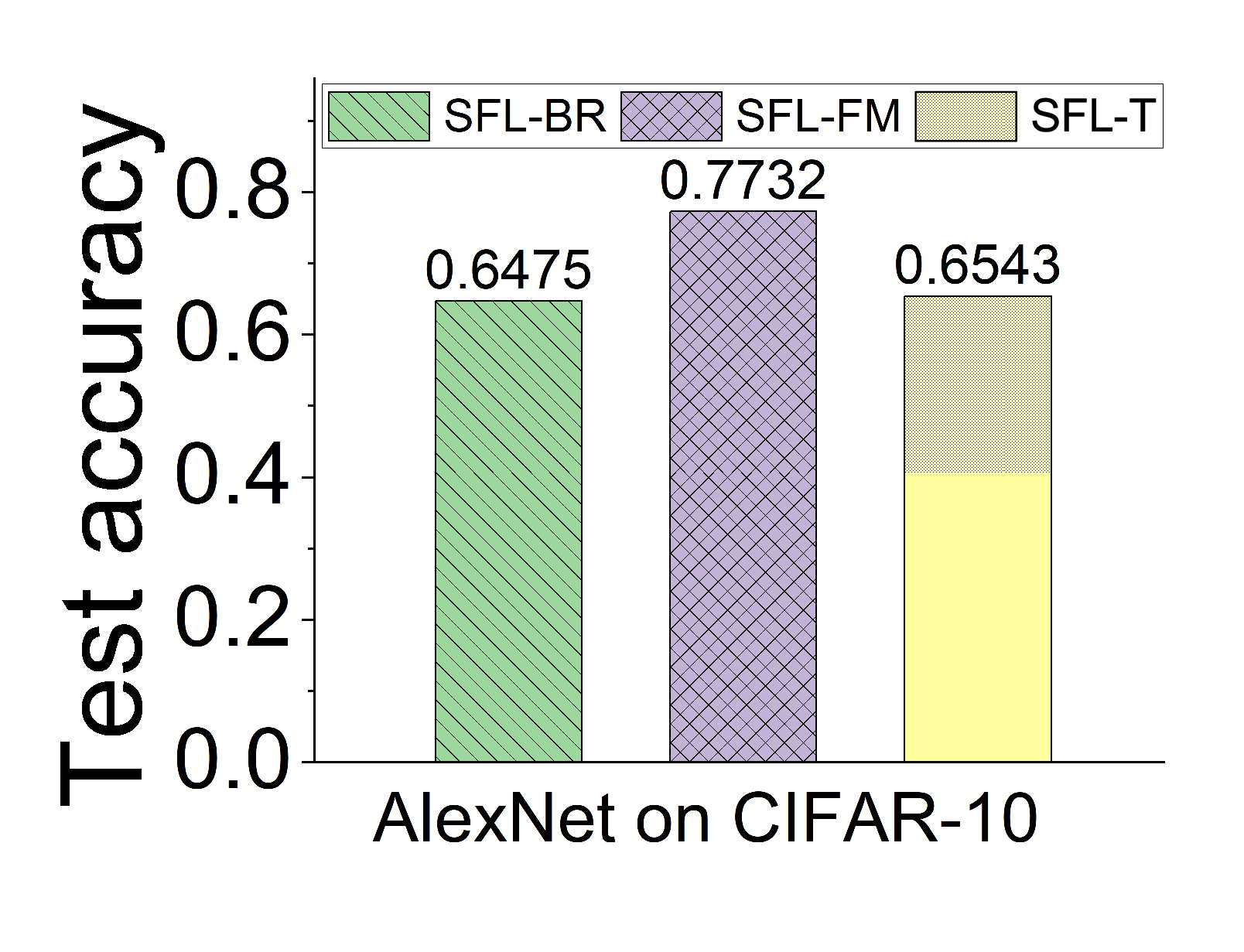}
    \label{fig:selection_acc}
}
\vspace{-0.1cm}
\caption{Training performance of three approaches with non-IID data.}
\label{fig:selection_record}
\vspace{-0.3cm}
\end{figure}

\begin{figure}[t]
\centering
\subfigure[Top Model]
{
    \includegraphics[width=0.425\linewidth,height=2.8cm]{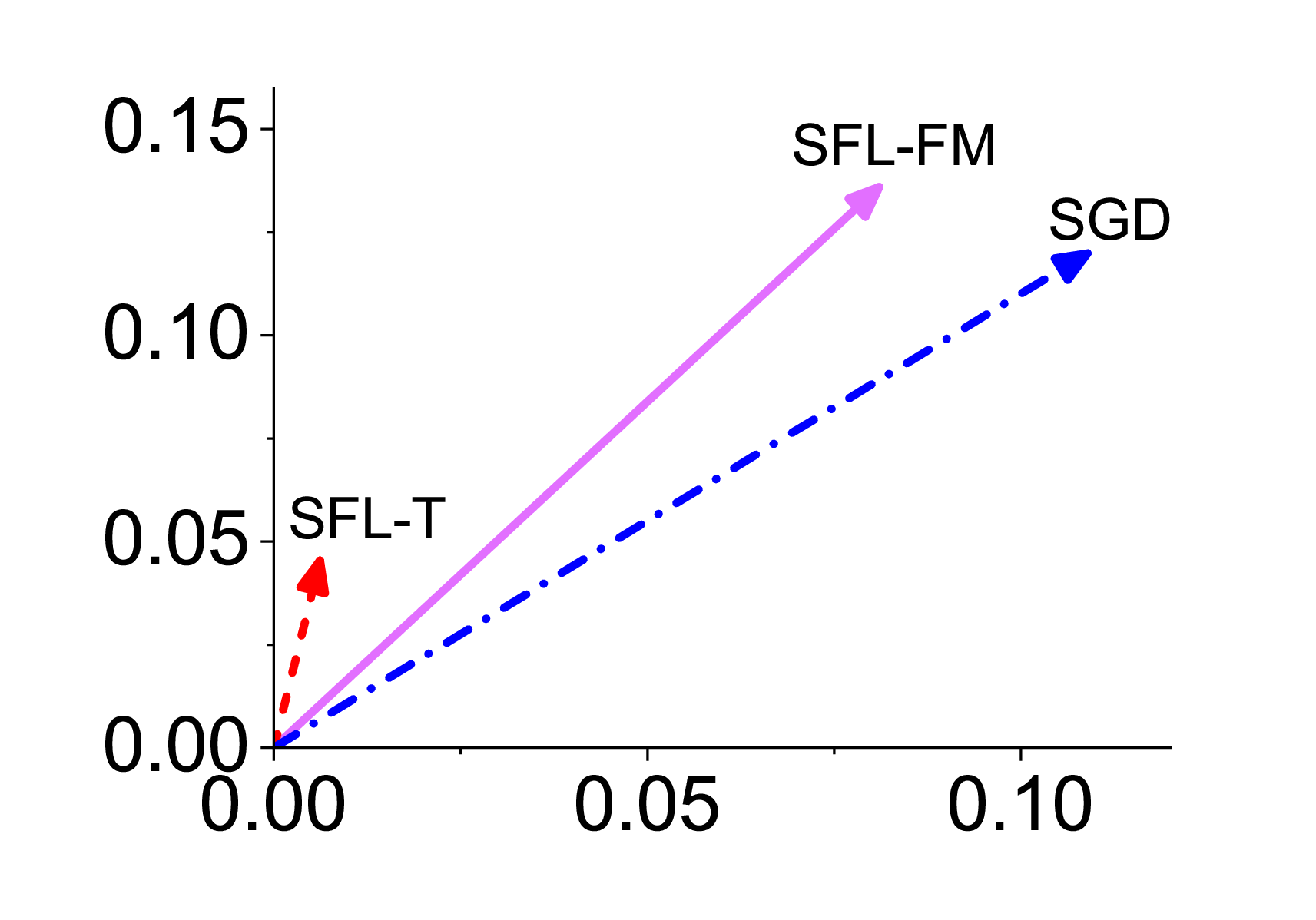}
    \label{fig:merging_server_vector}
}\quad 
\subfigure[Bottom Model]
{
    \includegraphics[width=0.425\linewidth,height=2.8cm]{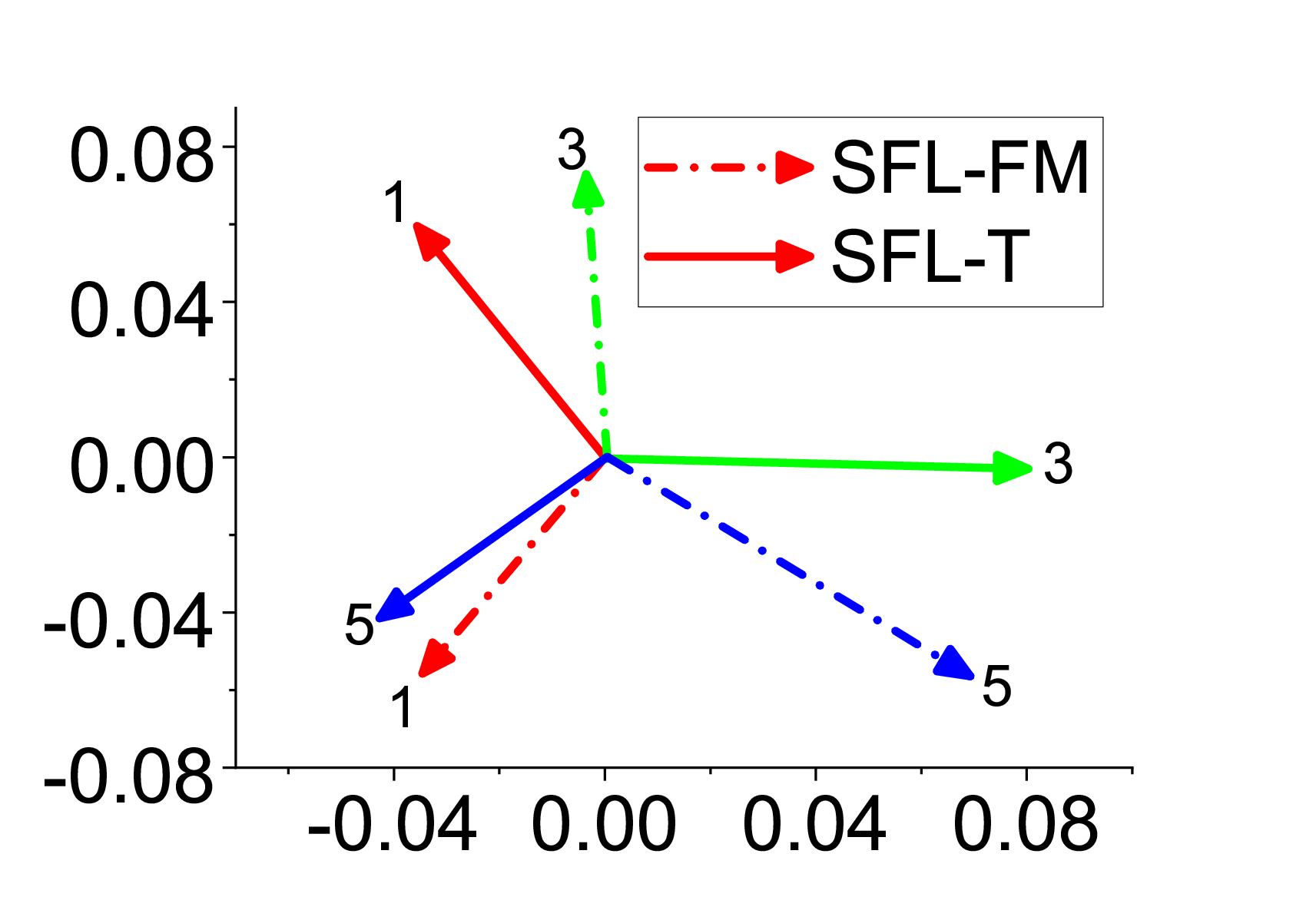}
    \label{fig:merging_workers_vector}
}
\vspace{-0.1cm}
\caption{Visualization of gradients of the top and bottom models.}
\label{fig:merging_vector}
\vspace{-0.5cm}
\end{figure}

To demonstrate the effectiveness of feature merging, we conduct a set of pre-experiments for training AlexNet on 10 workers with typical SFL (denoted as \textit{SFL-T} for short) and SFL with feature merging (denoted as \textit{SFL-FM}).
We distribute non-IID data samples from the CIFAR-10 dataset to the participating workers, whose data together are IID.
We record the training process and final test accuracy of models trained with SFL-T and SFL-FM.
By Figs. \ref{fig:selection_test} and \ref{fig:selection_record}, SFL-FM improves test accuracy by about 18.2\%, compared to SFL-T.
Furthermore, we decide to perform one iteration of model updating with SFL-FM and SFL-T, respectively, and investigate the effects of feature merging on gradients.
The iteration starts with the same top and bottom models, and the mini-batches across workers are non-IID, while the union of the mini-batches follows IID.
In Fig. \ref{fig:merging_server_vector}, the backpropagated gradients from the top models of SFL-T and SFL-FM are visualized in the 2D vector space by performing principal component analysis (PCA) \cite{abdi2010principal}.
Besides, we also conduct standalone model training of the entire model (\ie, a combination of the top and bottom models) for one iteration \wrt the above (IID) mini-batch union.
The gradient derived by the standalone SGD generally indicates the right optimization direction.
Moreover, the gradients corresponding to bottom models of three randomly selected workers are illustrated in Fig. \ref{fig:merging_workers_vector}, where the dashed arrows and solid arrows denote the gradient vectors in SFL-T and SFL-FM, respectively.

By Fig. \ref{fig:merging_server_vector}, we observe that the gradient derived by SFL-FM is much closer to that by the standalone SGD, since the role of feature merging in SFL-FM can be regarded as a regularization operation for the gradients, which ensures updating the top model along a much more right direction than SFL-T.
Besides, the dispatched gradients in Fig. \ref{fig:merging_workers_vector} also exhibit quite different directions compared to the gradients by SFL-T, and help to update the bottom models more efficiently as validated in Fig. \ref{fig:selection_time_acc}.
In a nutshell, SFL-FM enables faster convergence rate and higher test accuracy for the trained model than SFL-T, which demonstrates the advantages of feature merging in addressing statistical heterogeneity.

\begin{figure*}[t]
\centering
\includegraphics[width=0.65\linewidth]{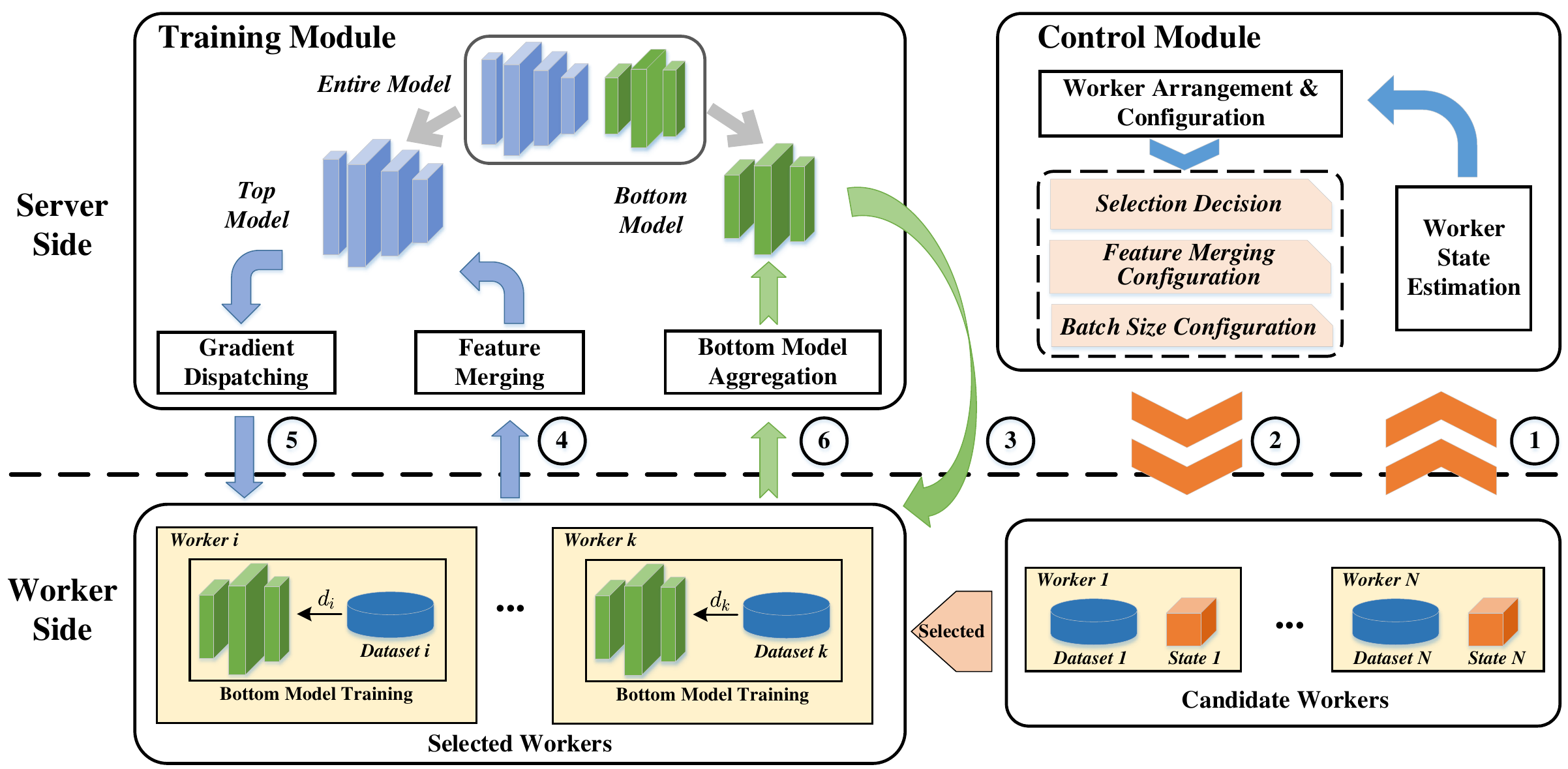}
\vspace{-0.2cm}
\caption{System workflow of \method. }
\label{fig:overview}
\vspace{-0.6cm}
\end{figure*}

\subsection{Importance of Batch Size Regulation}
Due to system heterogeneity, the computing time of bottom models and transmission time of features at each iteration probably vary significantly across workers.
If the workers are assigned with identical and fixed batch sizes at each iteration, the fast workers are forced to wait for the slow ones, incurring idle waiting time and inevitably impairing the training efficiency.
Accordingly, we propose to adaptively assign different batch sizes to workers with diverse capacities, termed \textit{batch size regulation}, so as to greatly reduce the waiting time and address the system heterogeneity.
Generally, the workers with higher computing and communication capabilities are configured with larger batch sizes, and can process more data at each iteration, while those with lower capabilities are assigned with smaller batch sizes.

To illustrate the efficiency of SFL with batch size regulation (denoted as \textit{SFL-BR}), we record the average waiting time and completion time of model training with SFL-BR and SFL-T.
As shown in Figs. \ref{fig:selection_test} and \ref{fig:selection_record}, with the help of batch size regulation, SFL-BR only takes 5,943s to achieve the target accuracy of about 65\%, while SFL-T takes 10,585s to reach the similar target accuracy.
Besides, by Fig. \ref{fig:selection_waiting_time}, SFL-BR reduces the average per-round waiting time by about 67\%, compared to SFL-T, which demonstrates its superiority in addressing system heterogeneity.

\vspace{-0.1cm}
\subsection{Discussion}
\vspace{-0.1cm}
Motivated by the above findings, it is necessary to incorporate feature merging and batch size regulation in SFL to simultaneously cope with system and statistical heterogeneity.
Given workers with diverse batch sizes matching their heterogeneous capacities, it is usually infeasible to directly merge the features from all workers for model training, since the underlying distribution of the mixed feature sequence probably deviates further away from that of the features derived from IID data.
An intuitive way is to select and arrange a part of workers, whose data together follows IID assumption, for feature merging.
However, there may still exist a large gap between the distribution of merged features from local data and the IID distribution.
Since batch size always imposes a comprehensive influence on the distribution of the merged features and computing/communication overhead, we are encouraged to reconfigure batch sizes for the selected workers, so as to 
contribute to well tackling the statistical and system heterogeneity.
Moreover, the inherent resource limitation in EC will complicate the optimization problem of SFL, which is elaborated in Section \ref{sec:design}.

\vspace{-0.2cm}
\section{System Overview}\label{sec:overview}
As illustrated in Fig. \ref{fig:overview}, \method consists of two key modules, \ie, the control module and the training module.
At the beginning of each round, the control module of the PS first collects the state information (\eg, label distribution, computing and communication capabilities) of all candidate workers (\circled{1}).
Subsequently, the control module estimates the worker states to dynamically select a part of workers with the consideration of statistical and system heterogeneity.
Once the control module has made the decision, it assigns the configurations of feature merging and batch size (\circled{2}) to the selected workers, which are further distributed with the bottom models (\circled{3}) and activated for model training.

In the training module, the selected workers train the bottom models using their local data and interact with the PS to update the top model at each iteration.
Concretely, the selected worker $i$ performs forward propagation with batch size of $d_i$, and then pushes the features (\circled{4}) to the PS.
At each iteration, the PS strives to obtain a large-size mixed feature sequence by merging the features from multiple workers to overcome the statistical heterogeneity.
Afterwards, the PS completes forward and backward propagation with the mixed feature sequence to update top model.
Following the backward propagation of the top model, the PS has to divide the mixed large-size gradients into small gradients corresponding to each worker.
Subsequently, the PS dispatches the corresponding gradients (\circled{5}) back to the selected workers.
After a certain number of local iterations, the PS aggregates the bottom models from all selected workers (\circled{6}) to get the updated bottom model for next-round training.
It is worth noting that, since the selected workers are configured with different batch sizes, the bottom models will be assigned with adaptive aggregation weights related to their batch sizes, so as to guarantee the convergence of bottom model when performing model aggregation.

\vspace{-0.2cm}
\section{System Design}\label{sec:design}
In this section, we will elaborate the detailed design of control and training modules in \method.

\vspace{-0.2cm}
\subsection{Control Module}
\vspace{-0.1cm}
In each communication round, the control module of the PS is implemented to estimate the state information of all available workers, and further generates specific configurations of feature merging and batch size for the workers upon their heterogeneous properties.
The workers that meet the selection requirements are arranged to participate in the model training.



\textbf{Worker State Estimation.} \label{sec:estimation}
In order to make effective decisions, it is crucial to collect information about the current working states of the PS (\eg, ingress bandwidth) as well as all workers (\eg, label distribution, computing and communication capabilities).

Concretely, in EC, the available ingress bandwidth $B^h$ of the PS is usually limited in each round $h$, and the PS always consumes a large portion of bandwidth to exchange features/gradients with the workers.
It is instinct to ensure that the occupied bandwidth of the PS does not exceed the bandwidth budget $B^h$ to prevent the PS from becoming a bottleneck.
Therefore, at the beginning of round $h$, estimating the available ingress bandwidth $B^h$ becomes vital to determine the number of selected workers and their corresponding batch sizes in \method.
We analyze the statistical distribution of the ingress bandwidth based on the behavior of the PS in the previous rounds, and employ the statistical results to estimate the available ingress bandwidth $B^h$ in round $h$.

To update the model along the relatively optimal direction and tackle the non-IID issue, \method merges the features from different workers to form a mixed feature sequence, which is approximately equivalent to the features derived from IID data.
Herein, the label distribution, a vector $\textbf{V}$ ($v_j \geq 0, j\in [1,M]$ and $\sum_{j=1}^{M}v_j=1$) to parameterize a categorical distribution of class labels over $M$ classes, is required to assist the implementation of feature merging.
As the workers will deliver the features with corresponding labels to continue forward propagation of the top model on the PS in typical SFL \cite{thapa2022splitfed, han2021accelerating, pal2021server}, the PS can directly collect the labels of workers' features and obtains the label distribution $\textbf{V}_i$ of worker $i$ \wrt the mini-batch.
However, if the workers are unwilling or not permitted to share labels, each worker $i$ will derive the label distribution $\textbf{V}_i$ based on its whole local data and report to the PS before training, which protects the label information of specific samples from be exposing.
Considering that privacy leakage is an important challenge in SFL, some popular privacy protection techniques, \eg, Differential Privacy \cite{thapa2021advancements, wu2023federated, ghazi2021deep}, Distance Correlation Technique \cite{vepakomma2019reducing} and Homomorphic Encryption \cite{yang2023dynamic}, can be applied to protect the privacy of raw data, features/gradients and models, and are orthogonal to the main focus of MergeSFL.

The estimation of time-varying computing and communication capacities of workers is critical for \method to develop appropriate strategies of batch size regulation for participating workers.
As proxy metrics, we adopt the computing time $\mu_i^h$ for processing one data sample and the corresponding transmission time $\beta_i^h$, which can be recorded by the workers directly during model training, to indicate the computing and communication capacities of worker $i$ in round $h$, respectively.
Prior to starting model training in round $h$, the PS collects the latest computing time $\hat{\mu}_i^h$ and transmission time $\hat{\beta}_i^h$ from all workers.
Besides, we introduce the moving average with the historical states of workers to improve the robustness of the estimation \cite{leroy2019federated}.
Accordingly, the PS estimates the computing time $\mu_i^{h}$ and corresponding transmission time $\beta_i^{h}$ for worker $i$ in round $h$ by calculating the moving average with $\alpha \in [0, 1]$ (for example, $\alpha=0.8$ in our experiments) as:
\vspace{-0.1cm}
\begin{equation}
\mu_i^{h}=\alpha \cdot \mu_i^{h-1}+ (1-\alpha) \cdot \hat{\mu}_i^h
\vspace{-0.1cm}
\end{equation}
\begin{equation}
\beta_i^{h}=\alpha \cdot \beta_i^{h-1}+ (1-\alpha) \cdot \hat{\beta}_i^h
\vspace{-0.1cm}
\end{equation}
Besides, it is worth noting that advancing the techniques for state estimation is not our focus, and other existing advanced estimation techniques \cite{halperin2010predictable, yue2017linkforecast} can be applied in \method.

\begin{algorithm}[!t]
\caption{Decision making for worker arrangement and configuration in round $h$.}\label{alg}
\begin{flushleft}
    {\bf Input:} $K_i$, $\textbf{V}_i$, $\mu_i^h$, $\beta_i^h$, $B^h$.
\end{flushleft}
\begin{algorithmic}[1]
    \STATE Assign the fastest worker $l$ with batch size $d_l^h =\mathcal{D}$.

    \STATE Calculate diverse batch sizes for all workers by Eq. \eqref{eq:waiting_constraint}. 

    \STATE Calculate the priority of all workers by Eq. \eqref{eq:selected_probability}. 

    \STATE Select $m$ workers based on their priority as the initial population and encode these workers.

    \STATE Perform the genetic algorithm to construct the worker set $\mathcal{S}^h$ to minimize $KL(\Phi_h || \Phi_0)$ under the resource constraint in Eq. \eqref{eq:bandwidth_constraint}.

    \STATE Formulate the minimization of $\Delta (\mathcal{S}^h)$ as a Lagrange dual problem and finetune the batch sizes of workers in $\mathcal{S}^h$ under the constraint of $KL(\Phi_h || \Phi_0) \le \varepsilon$. 

    \STATE Scale up or down the batch size proportionally to maximize the utilization of bandwidth resource without violating Eq. \eqref{eq:bandwidth_constraint}.
\end{algorithmic}
\begin{flushleft}
    {\bf Output:}
    The worker set $\mathcal{S}^h$ with specific configurations for training.
\end{flushleft}
\end{algorithm}

\textbf{Worker Arrangement and Configuration.} \label{sec:arrangement}
Based on the estimated state information, the control module tries to select and arrange a part of workers, which are dynamically configured with appropriate batch sizes and a feature merging plan.
The detailed process is presented in Alg. \ref{alg}.

We denote the local updating frequency as $\tau$, which is fixed and identical for workers during training as in typical SFL like LocFedMix-SL \cite{oh2022locfedmix}.
Therefore, given the estimated computing time $\mu_i^h$ and the corresponding transmission time $\beta_i^h$ of worker $i$, we formulate the duration time (including computing and communication time) of worker $i$ with batch size $d_i^h$ in round $h$ as follows:
\vspace{-0.2cm}
\begin{equation}
    t_i^h= \tau \cdot d_i^h \cdot (\mu_i^h + \beta_i^h)
    \vspace{-0.1cm}
\end{equation}
Therefore, when training with all workers, we can obtain the completion time of round $h$ upon the duration time of all workers as $t^h=\max\{t_i^h|\forall i \in [N]\}$, which equals to the duration time of the slowest worker.
Thus, the waiting time of worker $i$ in round $h$ can be expressed as $t^h-t_i^h$.
Accordingly, the average waiting time of all workers is formulated as:
\vspace{-0.2cm}
\begin{equation}
    \mathcal{W}^h =\frac{1}{N} \sum_{i=1}^{N}(t^h-t_i^h)
    \vspace{-0.1cm}
\end{equation}
To minimize the average waiting time, \method will regulate the batch sizes of all workers so as to align their duration time. 
Thus, it ensures that the average waiting time will be small enough to mitigate the negative impacts of synchronization barrier and improve the training efficiency.
The regulation rule is expressed as follows:
\vspace{-0.1cm}
\begin{equation}\label{eq:waiting_constraint}
    d_i^{h} = d_l^{h} \cdot  \lfloor \frac{(\mu_l^{h} + \beta_l^{h})}{(\mu_i^{h} + \beta_i^{h})} \rfloor \ \  \forall i \in [1,...,N]
    \vspace{-0.1cm}
\end{equation}
where $l$ denotes the index of the fastest worker assigned with the default maximum batch size $\mathcal{D}$ in round $h$.
According to Eq. \eqref{eq:waiting_constraint}, we can obtain the specific batch sizes for all workers in round $h$ (Line 1-2 of Alg. \ref{alg}).

Due to the constraint of available ingress bandwidth $B^h$ in round $h$, it is actually infeasible to allow all the workers to participate in training.
Therefore, \method selects $R^h$ workers and constructs the worker set $\mathcal{S}^h$ for feature merging and model training.
The occupied bandwidth of the PS communicating with the workers in $\mathcal{S}^h$ is limited as follows:
\begin{equation}\label{eq:bandwidth_constraint}
    \sum_{i \in \mathcal{S}^h} d_i^h \cdot c \le B^h
    \vspace{-0.1cm}
\end{equation}
where $c$ is a constant and denotes the bandwidth occupied by transmitting the feature of one data sample.

To tackle the non-IID issue, the mixed feature sequence needs to be approximately equivalent to the features derived from IID data.
We first define the IID distribution as $\Phi_0$.
If the data of all workers follows IID distribution, we can get $\Phi_0=\frac{1}{N}\sum_{i=1}^N \textbf{V}_i$, where $\textbf{V}_i$ is the label distribution of worker $i$.
Considering the worker set $\mathcal{S}^h$ with size of $R^h=|\mathcal{S}^h|$ in round $h$, the label distribution of data from workers in $\mathcal{S}^h$ is denoted as:
\vspace{-0.1cm}
\begin{equation}
    \Phi_h = \sum_{i\in \mathcal{S}^h} \frac{d_i^h\cdot \textbf{V}_i}{\sum\nolimits_{i \in \mathcal{S}^h} d_i^h}
    \vspace{-0.1cm}
\end{equation}
The mixed feature sequence of the worker set $\mathcal{S}^h$ for feature merging is expected to meet the requirement that its label distribution $\Phi_h$ is approximately consistent with the IID distribution $\Phi_0$.
We introduce the KL-divergence $KL(\Phi_h || \Phi_0)$ to measure the gap between $\Phi_h$ and $\Phi_0$ as follows \cite{hershey2007approximating, goldberger2003efficient}:
\begin{equation}\label{eq:KL}
    KL(\Phi_h || \Phi_0) = \sum_{j=1}^{M} \Phi_h(v_j) \text{log} \frac{\Phi_h(v_j)}{\Phi_0(v_j)}
    \vspace{-0.1cm}
\end{equation}
In order to balance the contribution of all workers to model training, we define the participating frequency (denoted as $K_i$) to keep track of the number of times that worker $i$ engages in training.
Then, the priority $p_i$ of selecting worker $i$ for future training is expressed as follows:
\begin{equation}\label{eq:selected_probability}
    p_i=\frac{\sum_{i=1}^N (K_i+1)}{K_i+1}
    \vspace{-0.1cm}
\end{equation}
which indicates that the workers with small participating frequencies will have a large priority to be selected.
We employ the genetic algorithm (GA) \cite{katoch2021review, gen2023genetic, de1988learning} to construct the worker set $\mathcal{S}^h$ with the minimum $KL(\Phi_h || \Phi_0)$ under the resource constraint in Eq. \eqref{eq:bandwidth_constraint}.
In particular, we select $m$ workers (\eg, $m=N/2$) based on their priority as the initial population, and 
encode each gene as whether the worker is selected or not (Line of 3-5 in Alg. \ref{alg}).

In practice, there may still exist a large gap between the label distribution of worker set $\mathcal{S}^h$ and the IID distribution. 
Thus, we need to continue regulating the batch sizes of workers in $\mathcal{S}^h$ to further minimize $KL(\Phi_h || \Phi_0)$ and ensure $KL(\Phi_h || \Phi_0) \le \varepsilon$, where $\varepsilon \ge 0$ is the predefined threshold (close the zero).
However, the regulation of batch size inevitably violates the Eq. \eqref{eq:waiting_constraint} and increases the average waiting time of the worker set $\mathcal{S}^h$.
The increased waiting time at each iteration is denoted as:
\vspace{-0.1cm}
\begin{equation}\label{eq:BS}
    \Delta (\mathcal{S}^h) = \frac{1}{R^h} \sum_{i \in \mathcal{S}^h} (\Delta d_i^h \cdot (\mu_i^h + \beta_i^h))
    \vspace{-0.2cm}
\end{equation}
where $\Delta d_i^h$ is the difference of batch size before and after batch size regulation for worker $i$.
We explore to finetune the batch size so as to minimize $\Delta (\mathcal{S}^h)$ under the constraint of $KL(\Phi_h || \Phi_0) \le \varepsilon$.
To this end, we formulate the above problem as a Lagrange dual problem, which can be well solved as in \cite{gao2009canonical,burshtein2009iterative} (Line 6 in Alg. \ref{alg}).
After that, we scale up or down the batch size proportionally to maximize the utilization of bandwidth resource under the constraint in Eq. \eqref{eq:bandwidth_constraint} (Line 7 in Alg. \ref{alg}).

\subsection{Training Module}\label{sec:smashed_data_merging}
After the control module generates the worker selection decision, it distributes the feature merging and batch size configurations to the selected workers, which are employed to guide the subsequent model training process. 
Besides, the PS broadcasts the latest bottom models to the selected workers and runs the training module.
The training module consists of four phases, \ie, bottom model training, feature merging, gradient dispatching and bottom model aggregation.

\textbf{Bottom Model Training.}
For a certain worker $i$ in round $h$, we adopt the mini-batch SGD algorithm with batch size $d_i^h$ to update the bottom model.
In order to guarantee model convergence, we formulate a rule to guide the setting of worker-specific local learning rate $\eta_i^h$, which is proportional to the batch size $d_i^h$ of each worker $i$ as suggested in \cite{ma2021adaptive}.
Accordingly, the process of updating the bottom model at iteration $k+1$ is expressed as:
\begin{equation} \label{eq:bottom_training}
\boldsymbol{w}_{b,i}^{h,k+1}=\boldsymbol{w}_{b,i}^{h,k}- \eta_i^h \cdot \widetilde{\nabla} F_{b,i}(\boldsymbol{w}_{b,i}^{h,k})
\end{equation}



\textbf{Feature Merging.}
Once worker $i$ performs forward propagation at iteration $k$ in round $h$, it delivers its features $g_i^{h,k}$ and the corresponding labels to the PS.
In terms of the feature merging configuration, the PS will merge the received features from the selected workers and obtain a mixed feature sequence, which is expected to be the features derived from an IID mini-batch.
The mixed feature sequence from workers (including from worker $i$ to worker $j$ in the worker set $\mathcal{S}^h$) is denoted as $G^{h,k}=[g_{i}^{h,k},...,g_{j}^{h,k}], |j-i|=R^h$.
Thus, the PS performs forward/backward propagation with the mixed feature sequence $G^{h,k}$ to update the top model at iteration $k$ in round $h$ as follows:
\vspace{-0.1cm}
\begin{equation} \label{eq:server_update}
\boldsymbol{w}_{p}^{h,k+1}=\boldsymbol{w}_{p}^{h,k} - \eta^h \cdot \widetilde{\nabla} F_{p}(\boldsymbol{w}_{p}^{h,k})
\vspace{-0.1cm}
\end{equation}
where $ \widetilde{\nabla} F_{p}(\boldsymbol{w}_{p}^{h,k})=\widetilde{\nabla} \ell(G^{h,k}; \boldsymbol{w}_{p}^{h,k})/(\sum_{i\in \mathcal{S}^h} d_i^h)$ and $\eta^h$ is the learning rate of top model in round $h$.
If the workers only delivers their features without labels, the PS would return the output logits of top model to the workers for calculating loss locally.
Then the workers send the loss back to the PS for calculating the gradients and completing the backward propagation.


\textbf{Gradient Dispatching.}
After performing backward propagation of the top model, the PS obtains the mixed backpropagated gradients $\widehat{G}^{h,k}$ at iteration $k$ in round $h$.
In order to correctly update the bottom models of different workers, it is necessary for the workers to obtain the gradients corresponding to their features uploaded at the feature merging phase.
Concretely, the PS first segments the mixed large-size gradients $\widehat{G}^{h,k}$ into multiple small-size gradients $[\hat{g}_{i}^{h,k},...,\hat{g}_{j}^{h,k}]$ for the selected workers, including from worker $i$ to worker $j$ in worker set $\mathcal{S}^h$.
Then, the PS dispatches the corresponding gradients to the selected workers for completing backward propagation according to Eq. \eqref{eq:bottom_training}.



\textbf{Bottom Model Aggregation.}
After performing totally $\tau$ iterations in round $h$, the selected workers in worker set $\mathcal{S}^h$ push their bottom models to the PS for central aggregation.
Considering the workers are configured with different batch sizes for model training, the bottom models are updated and trained with varying degrees, which needs adaptive weight aggregation to guarantee the performance of aggregated bottom model \cite{xu2022adaptive, li2022auto}.
Therefore, the PS aggregates the bottom models with adaptive weights related to the batch sizes of selected workers as follows:
\vspace{-0.2cm}
\begin{equation}
\boldsymbol{w}_{b}^{h}=\sum_{i\in \mathcal{S}^h}\frac{d_i^h \cdot \boldsymbol{w}_{b,i}^{h}}{\sum_{i\in \mathcal{S}^h} d_i^h}
\vspace{-0.1cm}
\end{equation}
The aggregated bottom model is stored in the PS and will be distributed to future selected workers to continue further training, or be combined with the top model to form a complete model used for AI tasks.

\section{Experiments and Evaluation}\label{sec:evaluation}
\begin{table}[!t]
\caption{Device technical specifications.}
\label{table:jetson}
\centering
\begin{tabular}{lcc}
\hline
    & \textbf{AI Performance} & \textbf{GPU Type}  \\ 
\hline
Jetson TX2 & 1.33 TFLOPs & 256-core Pascal \\ 
Jetson NX  & 21 TOPs & 384-core Volta\\ 
Jetson AGX  & 32 TOPs & 512-core Volta\\ \hline \hline
& \textbf{CPU Type} & \textbf{ROM} \\  \hline
Jetson TX2 & Denver 2 and ARM 4 & 8 GB LPDDR4\\ 
Jetson NX & 6-core Carmel ARM 8 & 8 GB LPDDR4x\\ 
Jetson AGX  & 8-core Carmel ARM 8 & 32 GB LPDDR4x \\  
\hline
\end{tabular}
\vspace{-0.4cm}
\end{table}

\subsection{Experimental Settings}
\vspace{-0.1cm}

\textbf{System Deployment.}
We conduct extensive experiments to evaluate the performance of \method on an edge computing hardware prototype system.
Specifically, we employ a deep learning GPU workstation as the PS, which is equipped with an Intel(R) Core(TM) i9-10900X CPU, four NVIDIA GeForce RTX 2080Ti GPUs and 256 GB RAM.
In addition, we specify 80 NVIDIA Jetson kits,
including 30 Jetson TX2 devices, 40 Jetson NX devices, and 10 Jetson AGX devices, as workers to construct a heterogeneous system. 
The detailed technical specifications of Jetson TX2, NX and AGX are listed in Table \ref{table:jetson}.
Notably, the TX2 showcases a 256-core Pascal GPU and a CPU cluster consisting of a 2-core Denver2 and a 4-core ARM CortexA57.
The NX is outfitted with a 384-core NVIDIA Volta GPU and a 6-core NVIDIA Carmel ARMv8.2 CPU.
Jetson Xavier NX dramatically enhances the NVIDIA software stack over 10$\times$ the performance of Jetson TX2.
Lastly, the AGX stands out a 512-core NVIDIA Volta GPU and an 8-core NVIDIA Carmel ARMv8.2 CPU.

In the experiments, we build the software platform based on Docker Swarm \cite{merkel2014docker,naik2016building} and the PyTorch deep learning library \cite{paszke2019pytorch}.
The Docker Swarm, a distributed software development kit, facilitates the construction of a distributed system and enables the monitoring of each device's operational status.
The PyTorch library facilitates the implementation of model training on devices. 
Additionally, to streamline communication among devices, we implement MPI (Message Passing Interface) \cite{gabriel2004open}, which includes a collection of sending and receiving functions.
The experimental source code is available at \url{https://github.com/ymliao98/MergeSFL}.

\begin{figure*}[t]
\centering
\subfigure[HAR]
{
    \includegraphics[width=0.22\linewidth,height=3.2cm]{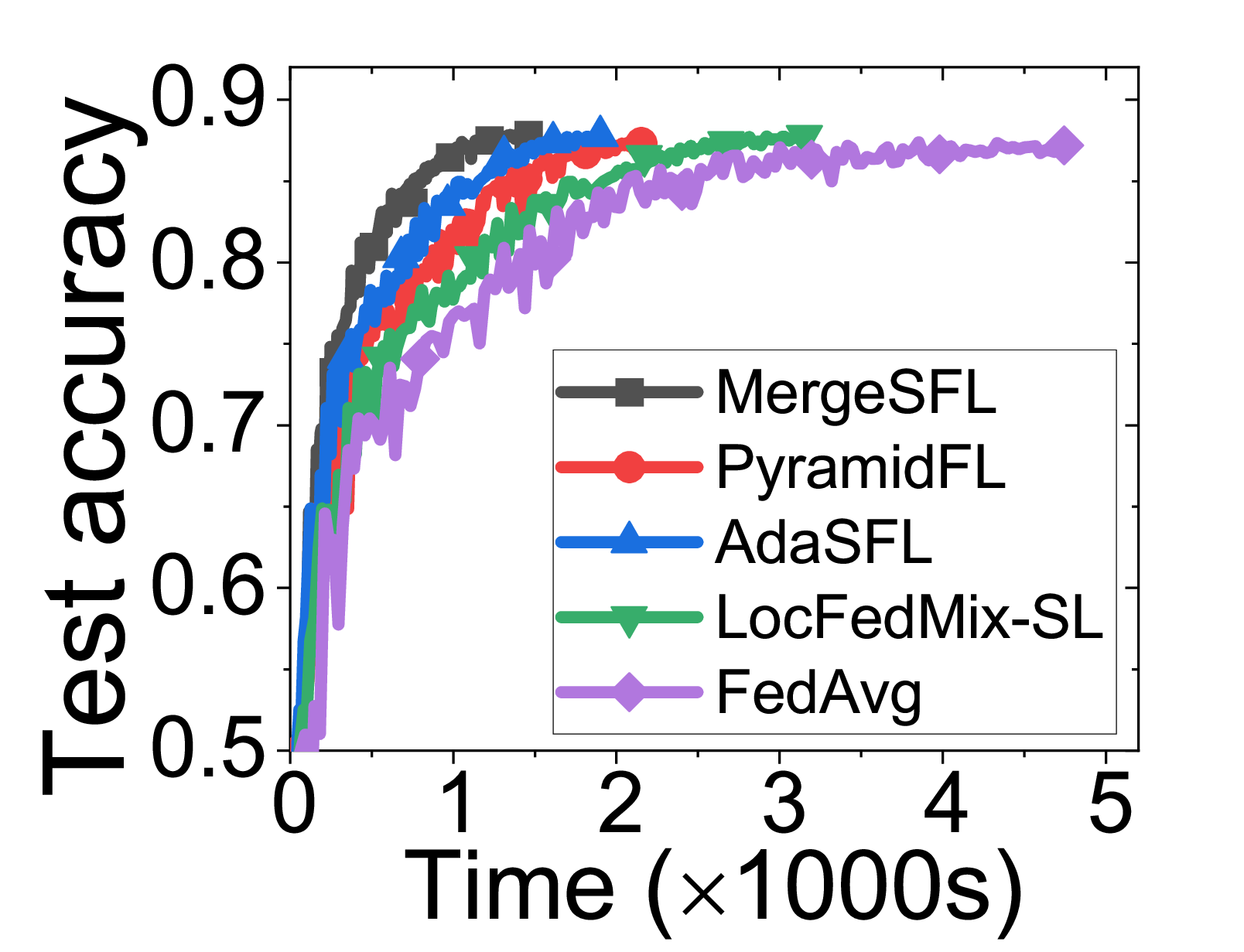}
    \label{fig:HAR-IID}
}\quad
\subfigure[Speech]
{
    \includegraphics[width=0.22\linewidth,height=3.2cm]{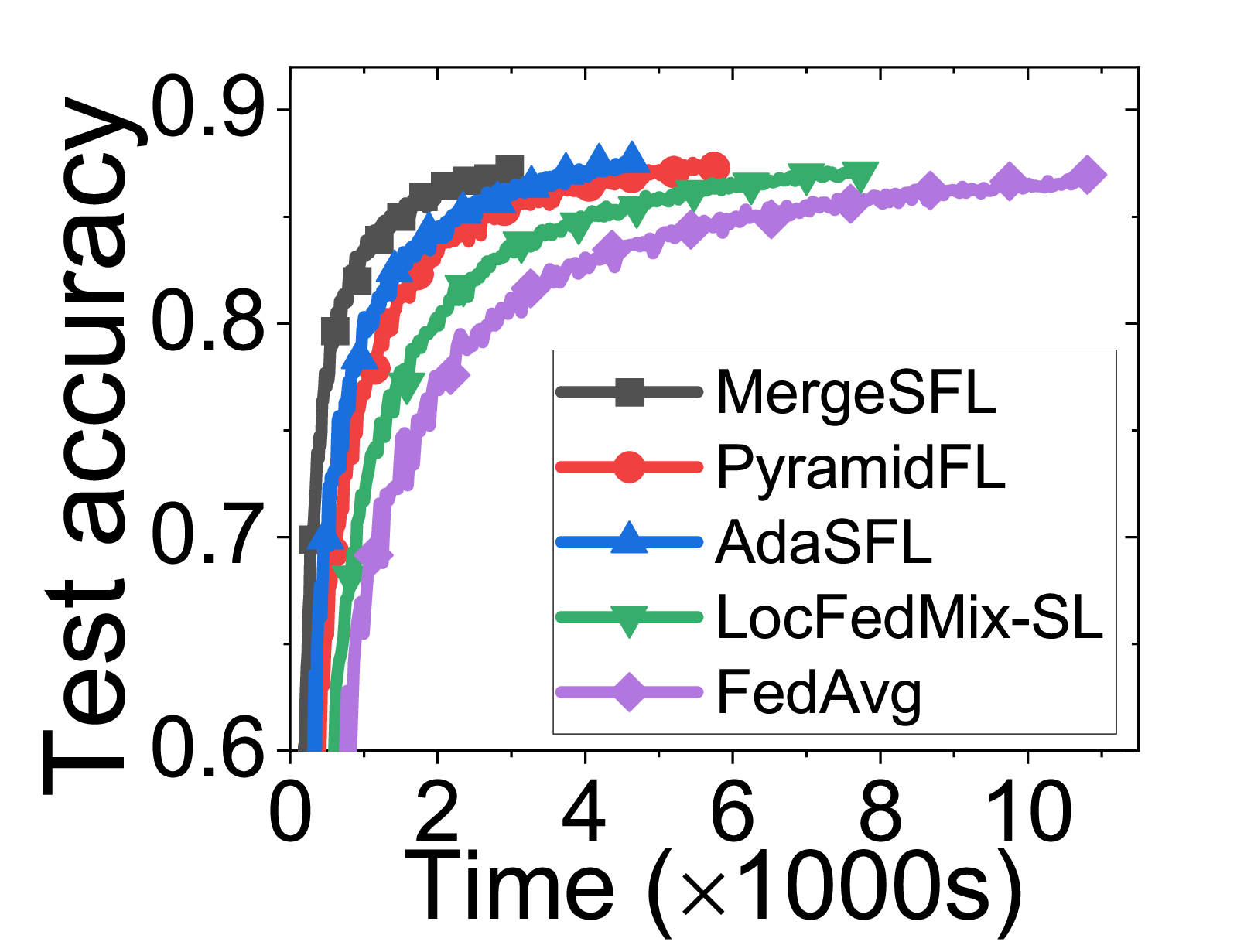}
    \label{fig:Speech-IID}
}\quad 
\subfigure[CIFAR-10]
{
    \includegraphics[width=0.22\linewidth,height=3.2cm]{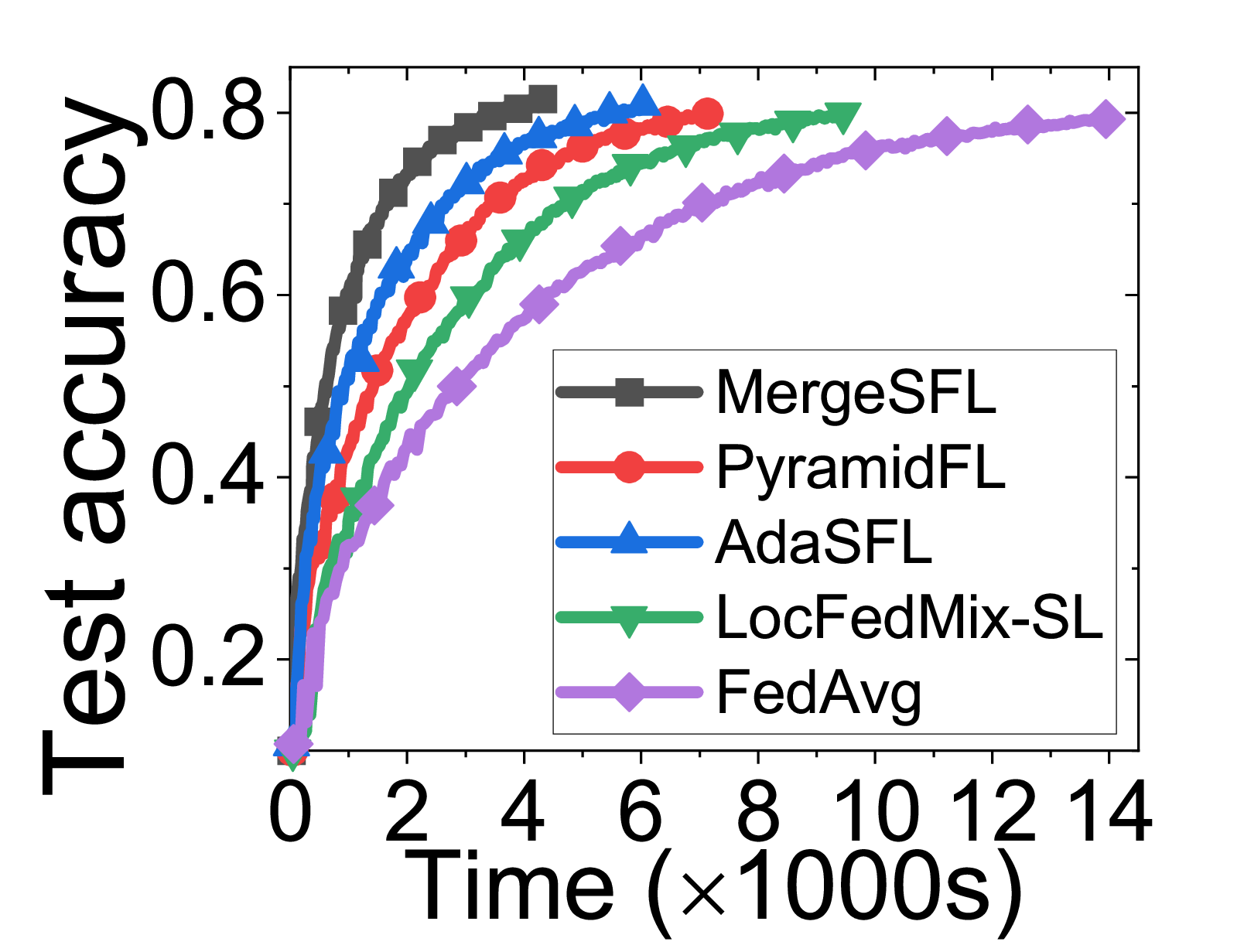}
    \label{fig:CIFAR10-IID}
}\quad 
\subfigure[IMAGE-100]
{
    \includegraphics[width=0.22\linewidth,height=3.2cm]{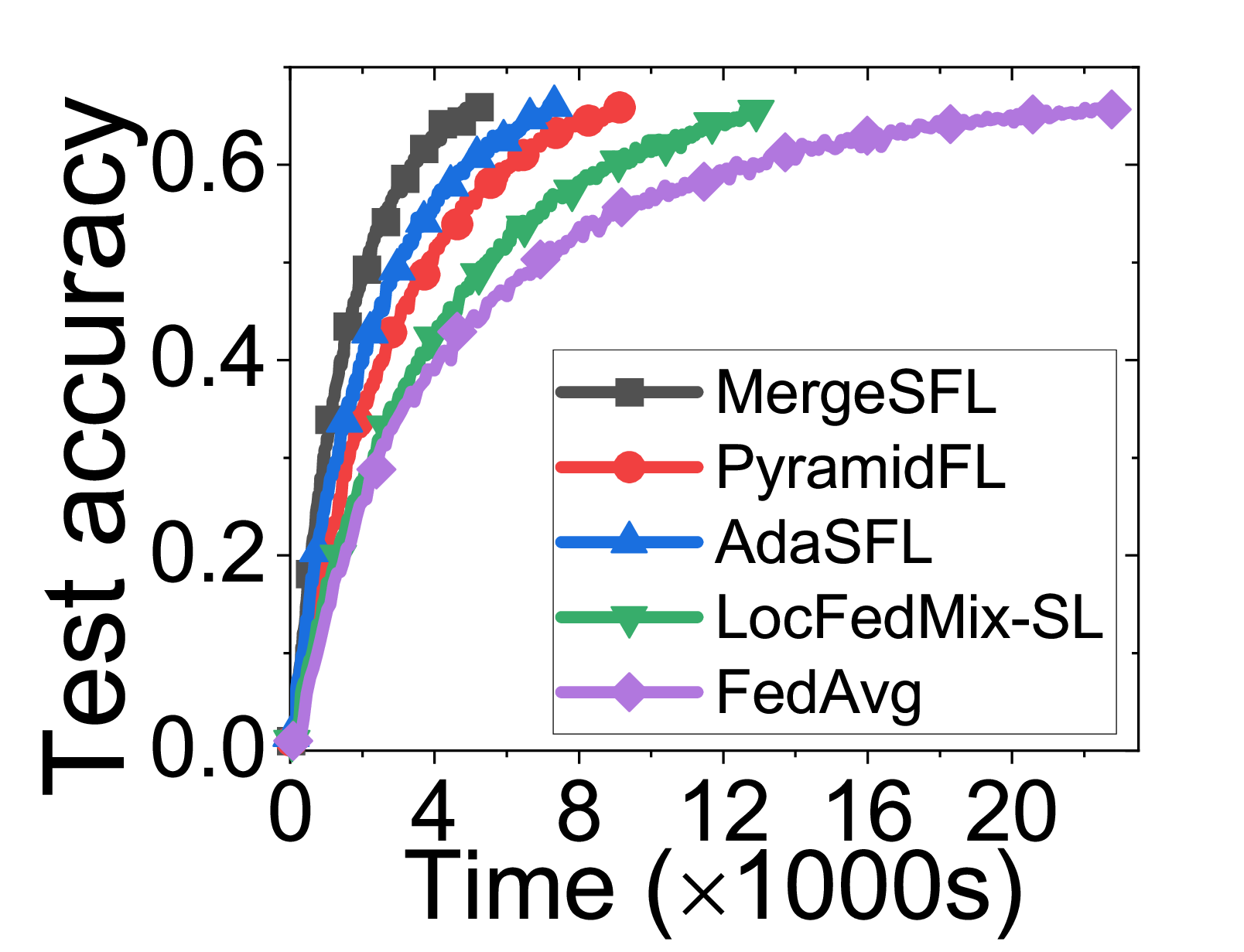}
    \label{fig:IMAGE100-IID}
}
\vspace{-0.2cm}
\caption{Test accuracy of five approaches on the four IID datasets.}
\label{fig:IID}
\vspace{-0.3cm}
\end{figure*}

\begin{figure*}[!t]
\centering
\subfigure[HAR]
{
    \includegraphics[width=0.22\linewidth,height=3.2cm]{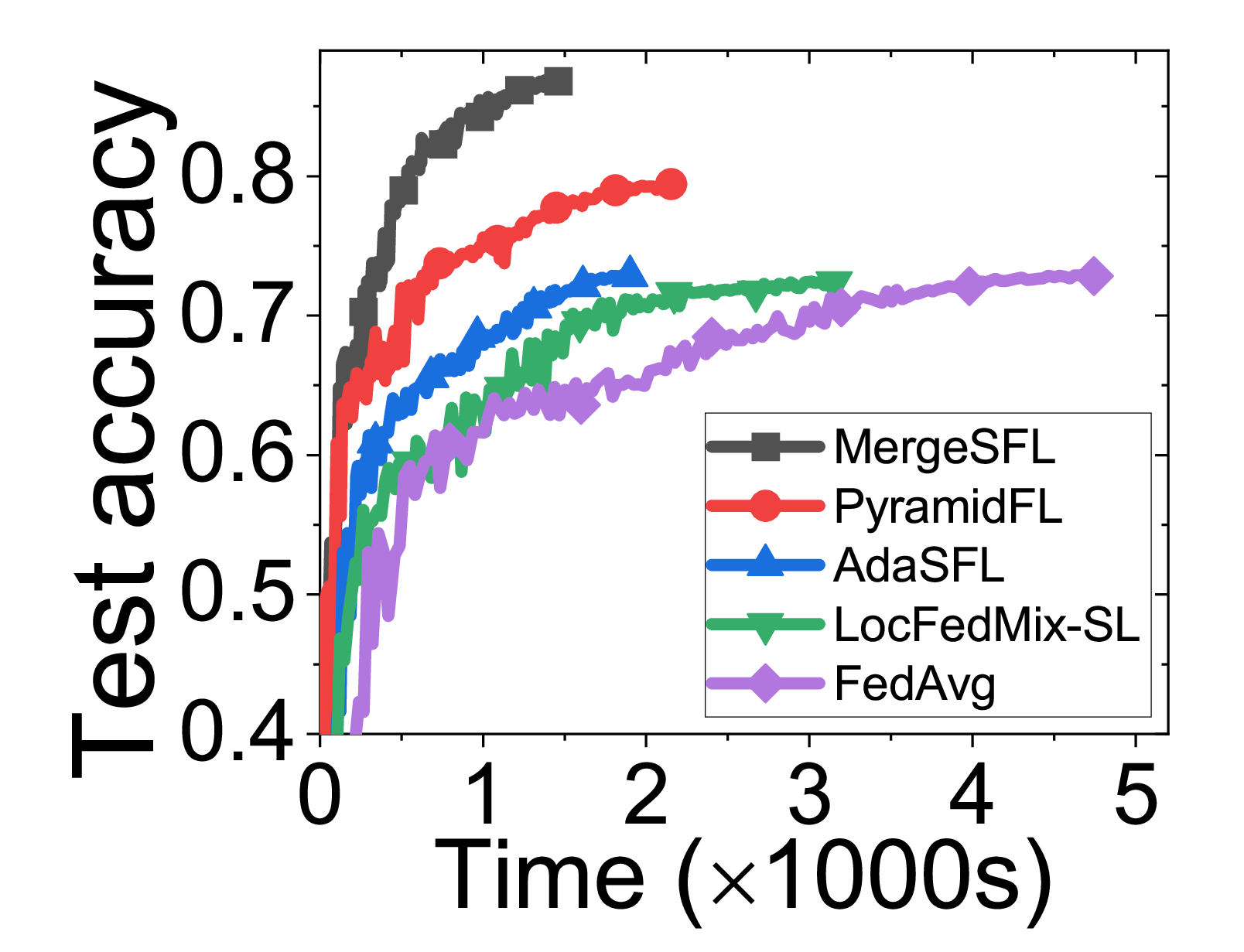}
    \label{fig:HAR-non-IID}
}\quad 
\subfigure[Speech]
{
    \includegraphics[width=0.22\linewidth,height=3.2cm]{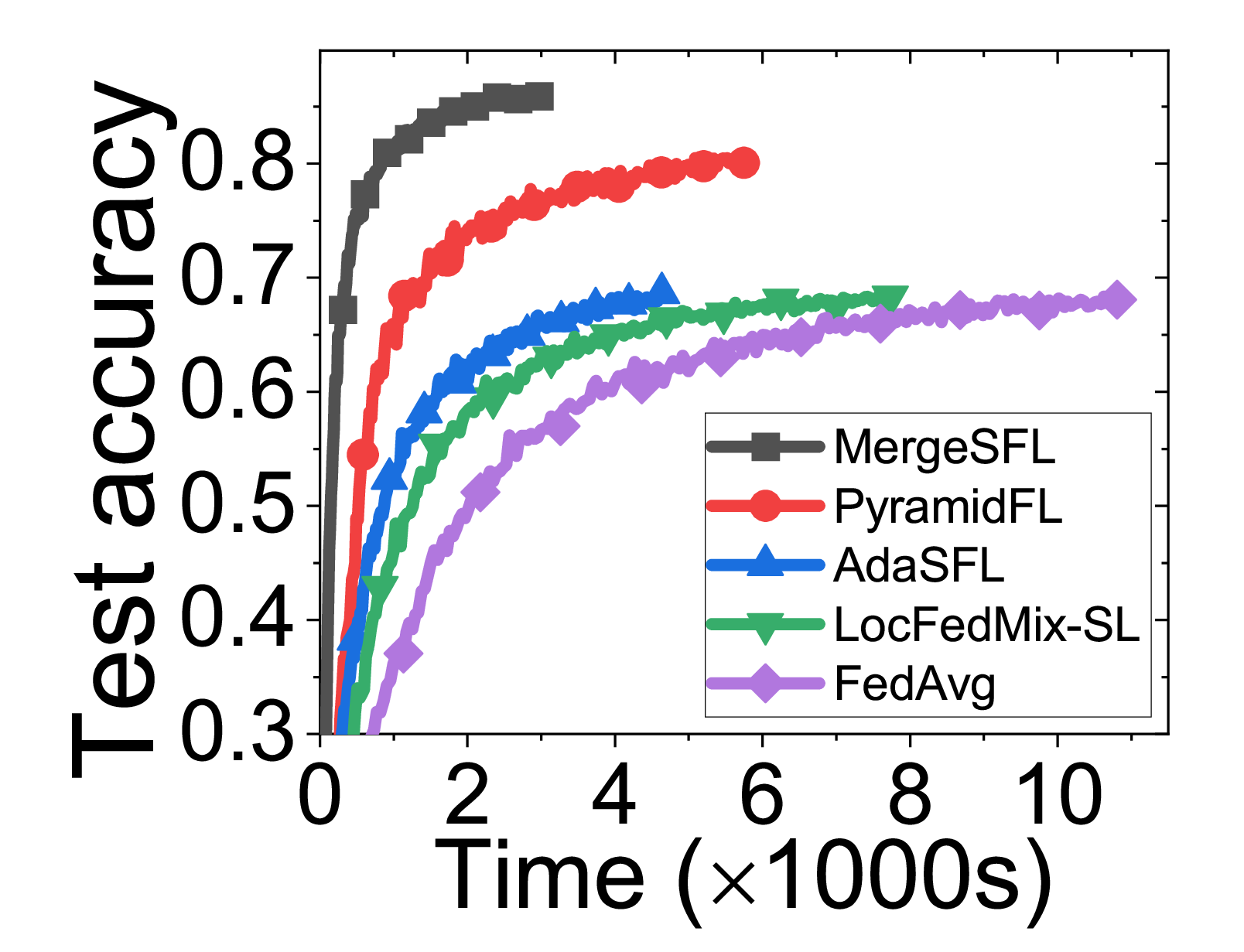}
    \label{fig:Speech-non-IID}
}\quad 
\subfigure[CIFAR-10]
{
    \includegraphics[width=0.22\linewidth,height=3.2cm]{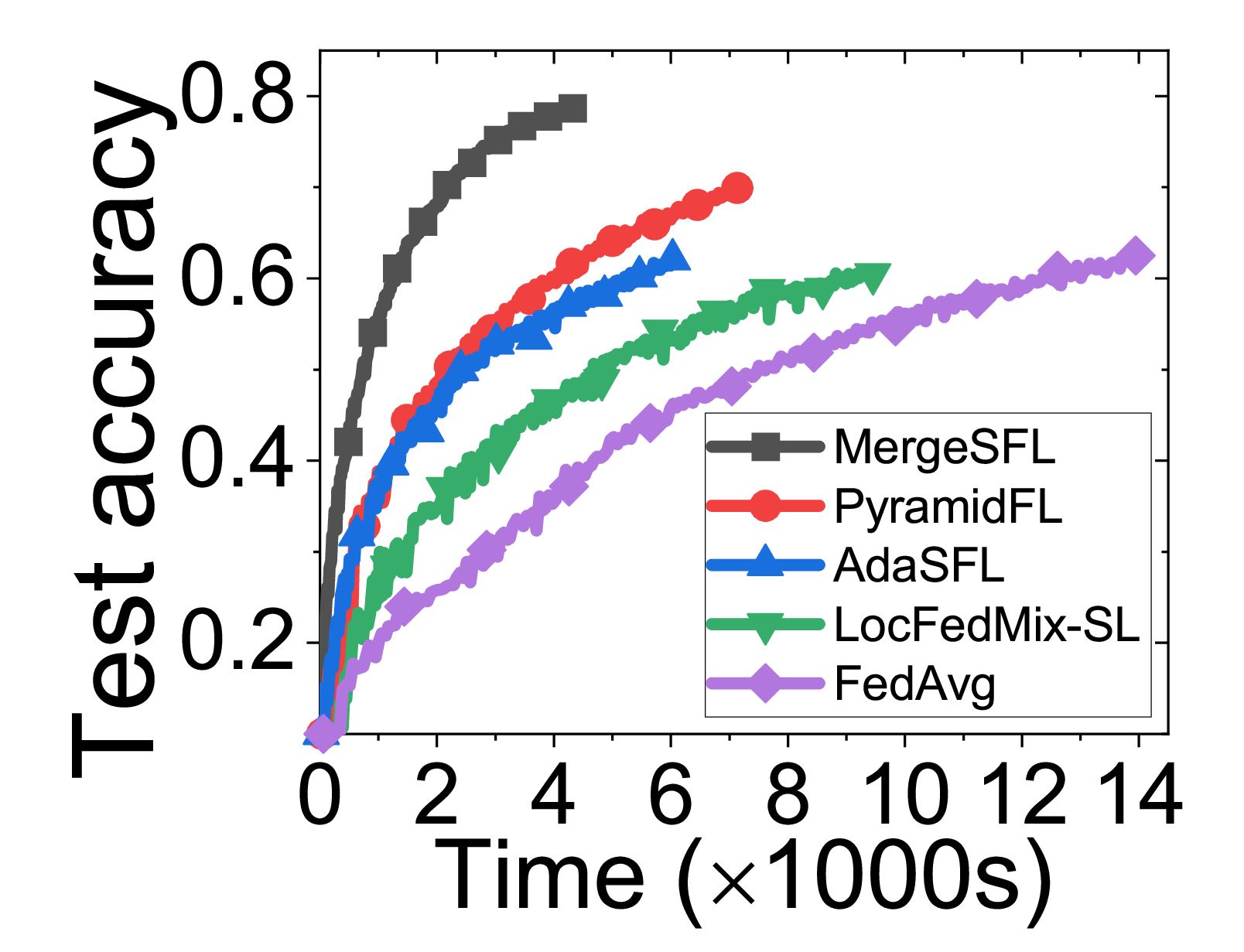}
    \label{fig:CIFAR10-non-IID}
}\quad 
\subfigure[IMAGE-100]
{
    \includegraphics[width=0.22\linewidth,height=3.2cm]{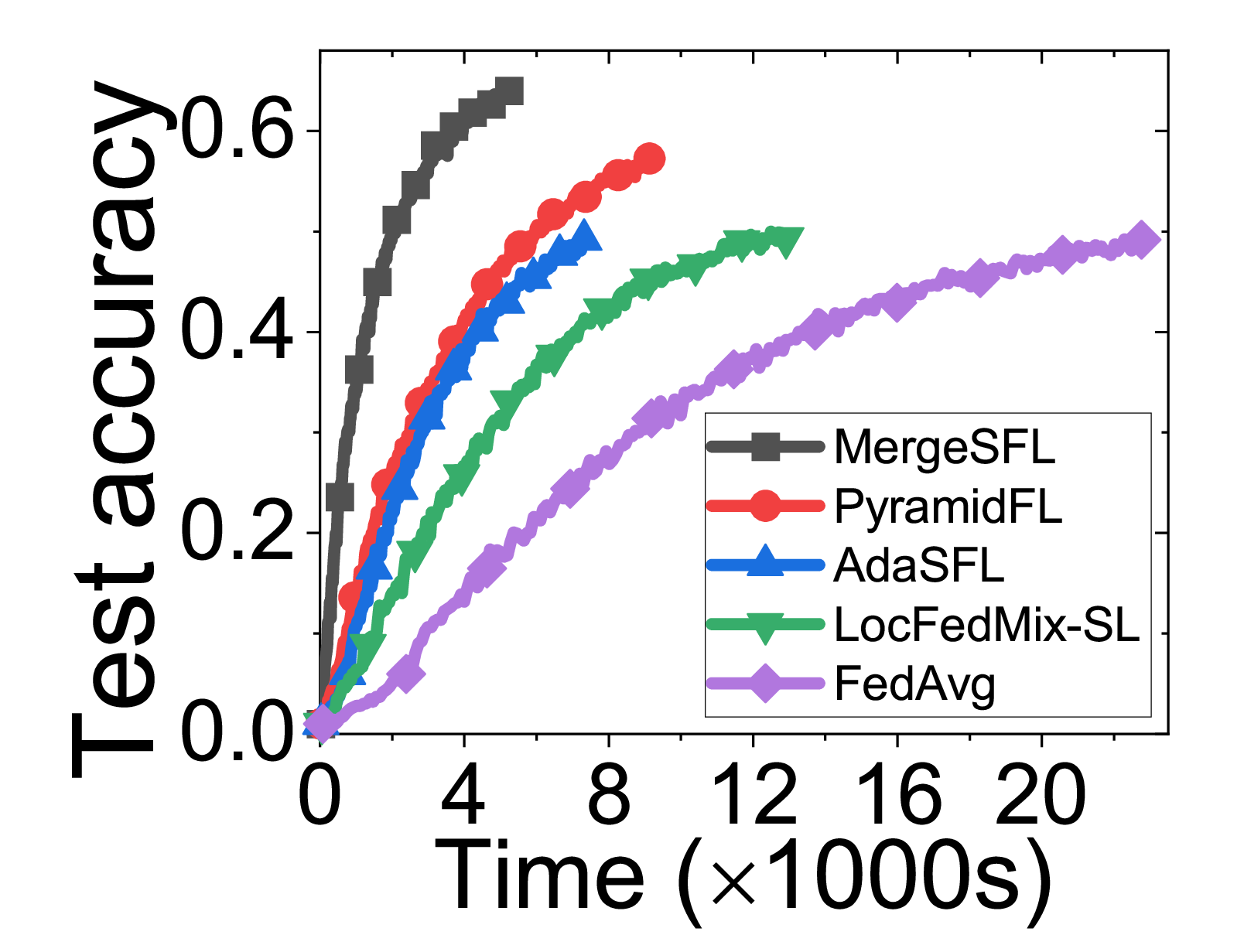}
    \label{fig:IMAGE100-non-IID}
}
\vspace{-0.2cm}
\caption{Test accuracy of five approaches on the four non-IID datasets.}
\label{fig:non-IID}
\vspace{-0.5cm}
\end{figure*}

\textbf{Setting of System Heterogeneity.}
To enable the workers with heterogeneous computing and communication capabilities, we present the following experimental settings.

1) \textbf{\textit{For Computation.}}
All the Jetson TX2, NX and AGX can be configured to work with different modes, specifying the number of working CPUs and the frequency of CPU/GPU, so that they can work with different computing capacities.
Specifically, TX2 can work in one of four modes while NX and AGX work in eight modes each.
For instance, the AGX with highest performance mode (\ie, mode 0 of AGX) achieves training by 100$\times$ faster than the TX2 with lowest performance mode (\ie, mode 1 of TX2).
To further reflect the time-varying on-device resources, we randomly change the modes for devices every 20 communication rounds.

2) \textbf{\textit{For Communication.}}
All devices are connected to the PS via WiFi routers.
We group the devices into four groups, each containing 20 devices.
These groups are then placed at different locations, \ie, 2m, 8m, 14m, and 20m away from the WiFi routers.
Due to random channel noise and competition among devices, the bandwidth between the PS and devices dynamically varies during the training.
The bandwidth of devices is measured by iperf3 \cite{tirumala1999iperf}, which fluctuates between 1Mb/s and 30Mb/s.

\textbf{Applications and Models.}
We evaluate the performance of \method on four classical datasets and four DNN models.

1) \textbf{\textit{Human Activity Recognition.}} 
We adopt the Human Activity Recognition (HAR) dataset \cite{anguita2013public} in this application, which is collected from 30 individuals and includes 7,352 samples for training and 2,947 for test.
We train a plain CNN model \cite{mcmahan2017communication} with three 5$\times$5 convolutional layers and two fully-connected layers, which is tailored to the HAR dataset and represented as CNN-H.

2) \textbf{\textit{Speech Recognition.}}
The Google Speech dataset \cite{warden2018speech} (expressed as Speech for short) is adopted for the task of speech recognition, which allows a computer or device to recognize and interpret spoken language.
The dataset includes 85,511 and 4,890 audio clips for training and test, respectively.
The model trained on Speech is a CNN network (denoted as CNN-S) with four 1-D convolutional layers and one fully-connected layer.

3) \textbf{\textit{Object Recognition.}}
We adopt the CIFAR-10 dataset \cite{krizhevsky2010convolutional} for the evaluation, which is an image dataset composed of 60,000 32$\times$32 colour images (50,000 for training and 10000 for test) across 10 categories.
We utilize an 8-layer AlexNet with size of 136MB \cite{krizhevsky2012imagenet} for CIFAR-10.
The AlexNet is composed of three 3$\times$3 convolutional layers, one 7$\times$7 convolutional layer, one 11$\times$11 convolutional layer, two fully-connected hidden layers, and one softmax output layer.

\begin{figure*}[!t]
\centering
\subfigure[HAR]
{
    \includegraphics[width=0.22\linewidth,height=3.2cm]{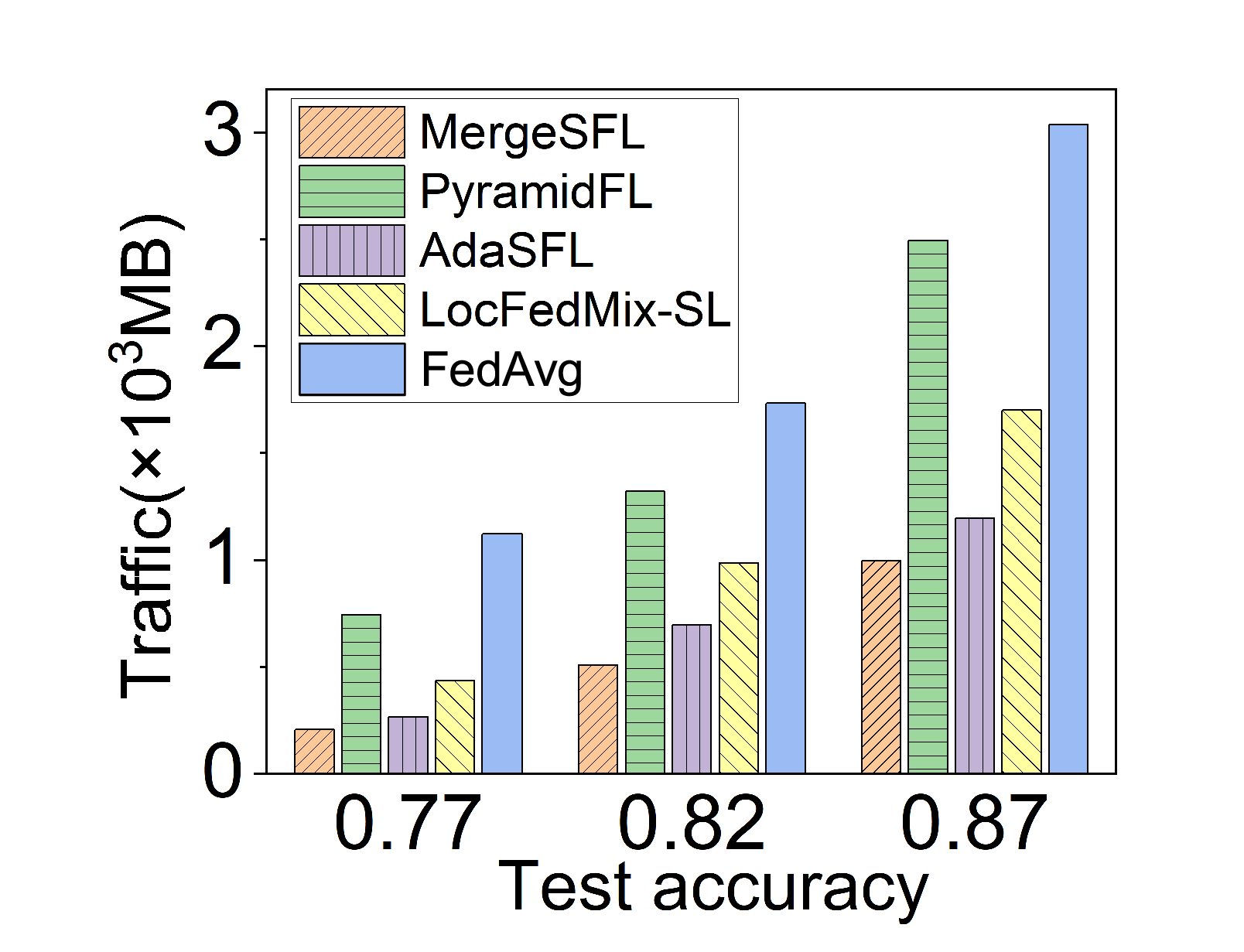}
    \label{fig:bandwidth_HAR}
}\quad 
\subfigure[Speech]
{
    \includegraphics[width=0.22\linewidth,height=3.2cm]{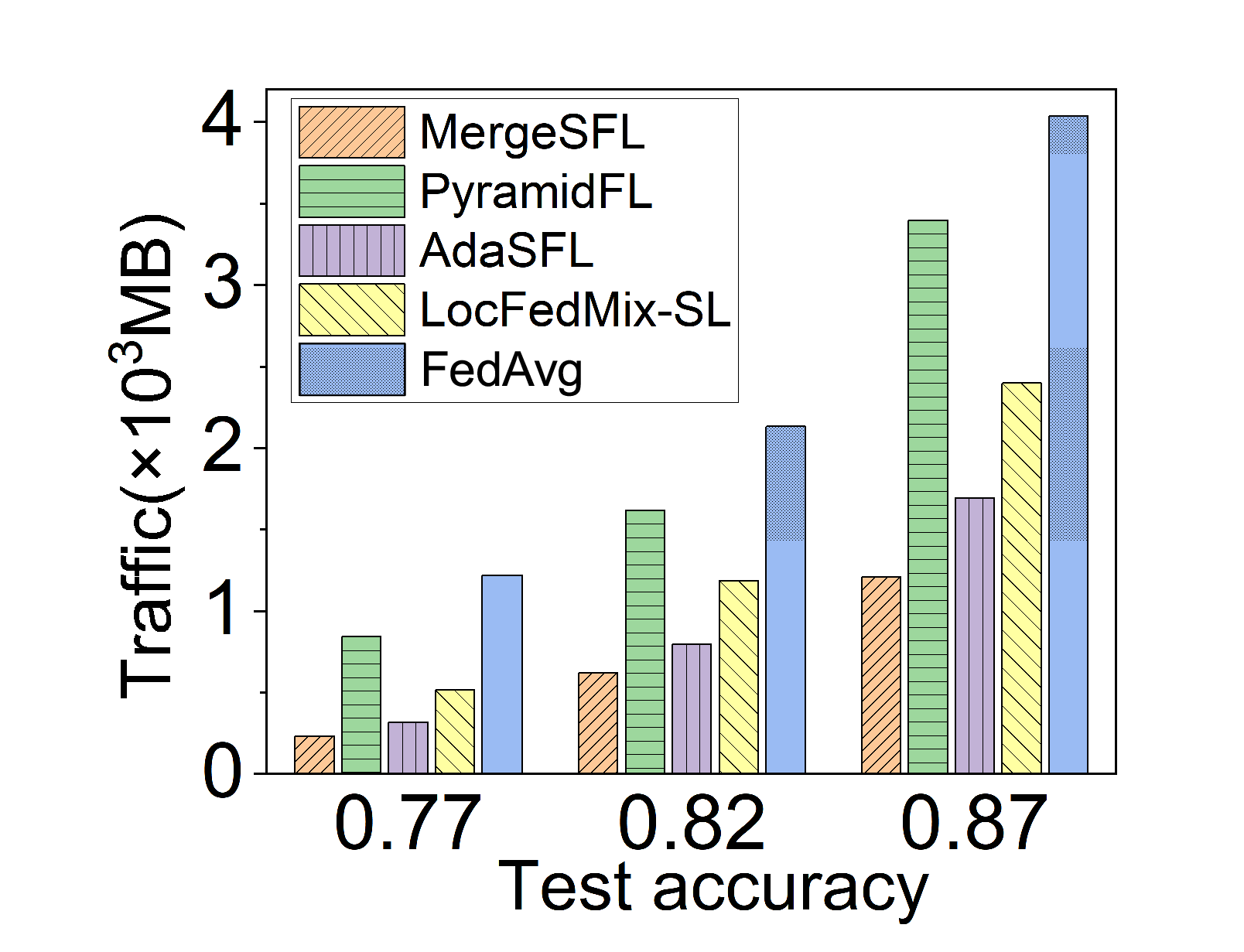}
    \label{fig:bandwidth_speech}
}\quad 
\subfigure[CIFAR-10]
{
    \includegraphics[width=0.22\linewidth,height=3.2cm]{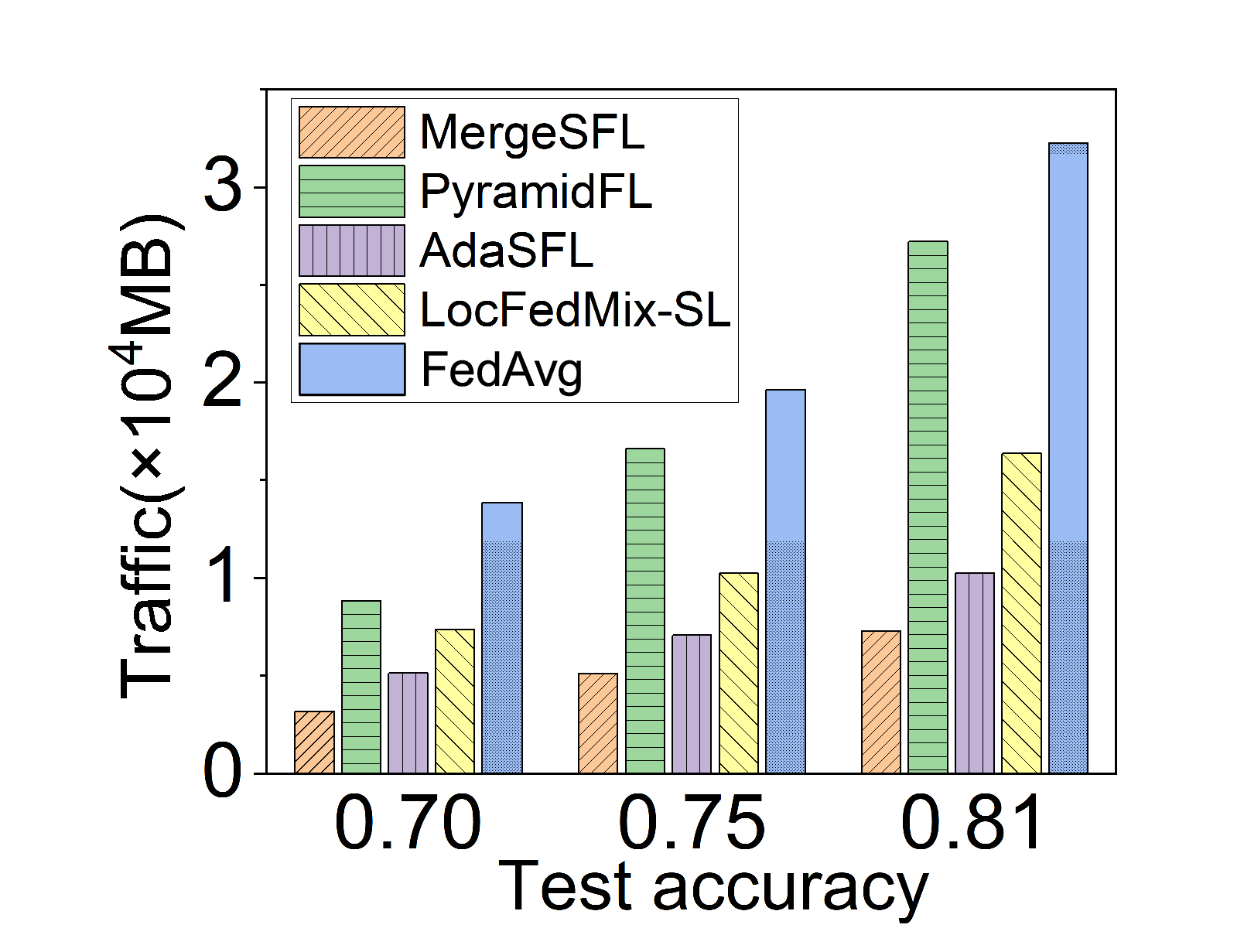}
    \label{fig:bandwidth_CIFAR10}
}\quad 
\subfigure[IMAGE-100]
{
    \includegraphics[width=0.22\linewidth,height=3.2cm]{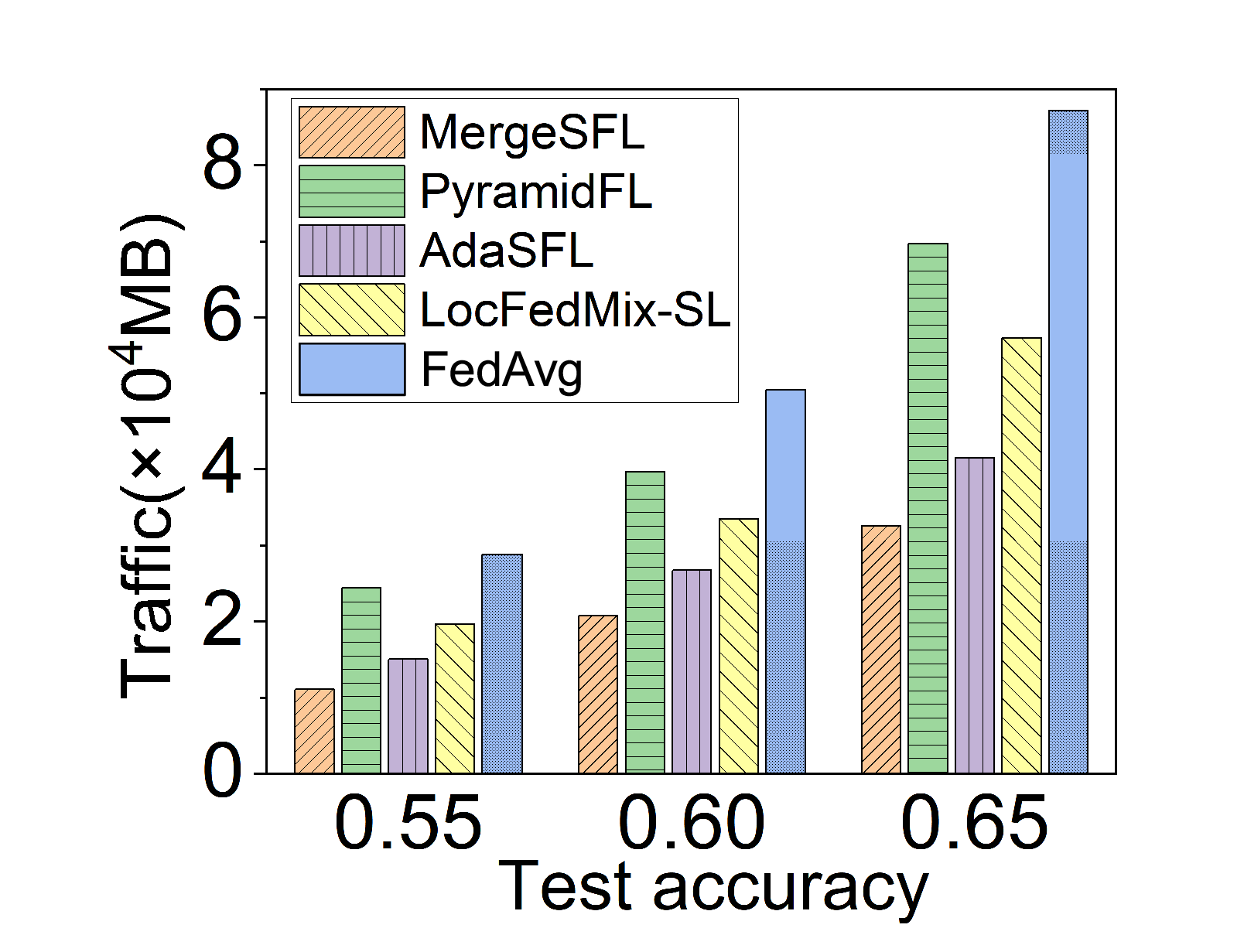}
    \label{fig:bandwidth_image}
}
\vspace{-0.2cm}
\caption{Network traffic consumption of five approaches when achieving different target accuracies.}
\label{fig:bandwidth}
\vspace{-0.2cm}
\end{figure*}

\begin{figure*}[t]
\centering
\subfigure[HAR]
{
    \includegraphics[width=0.22\linewidth,height=3.2cm]{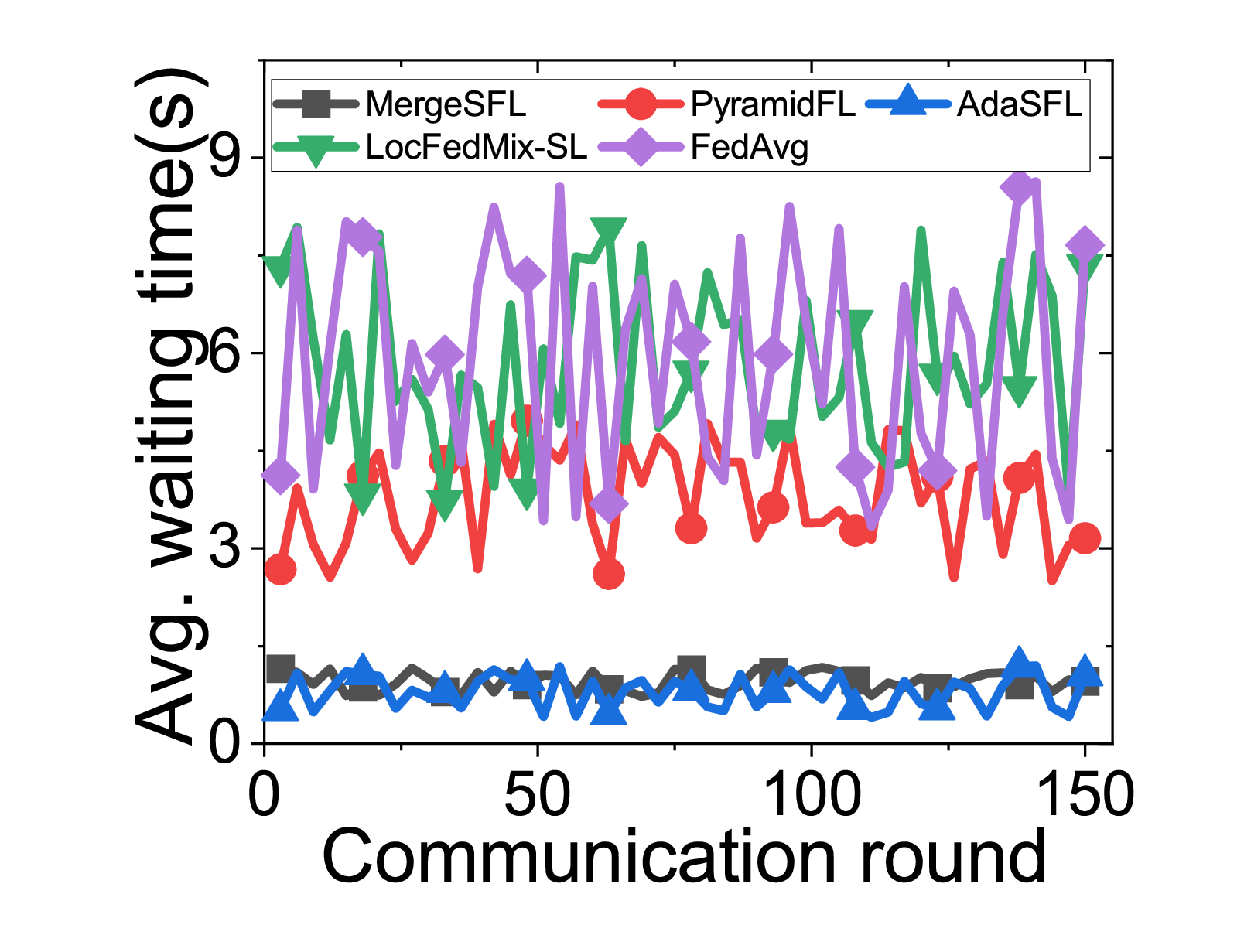}
    \label{fig:HAR-waiting_time}
}\quad 
\subfigure[Speech]
{
    \includegraphics[width=0.22\linewidth,height=3.2cm]{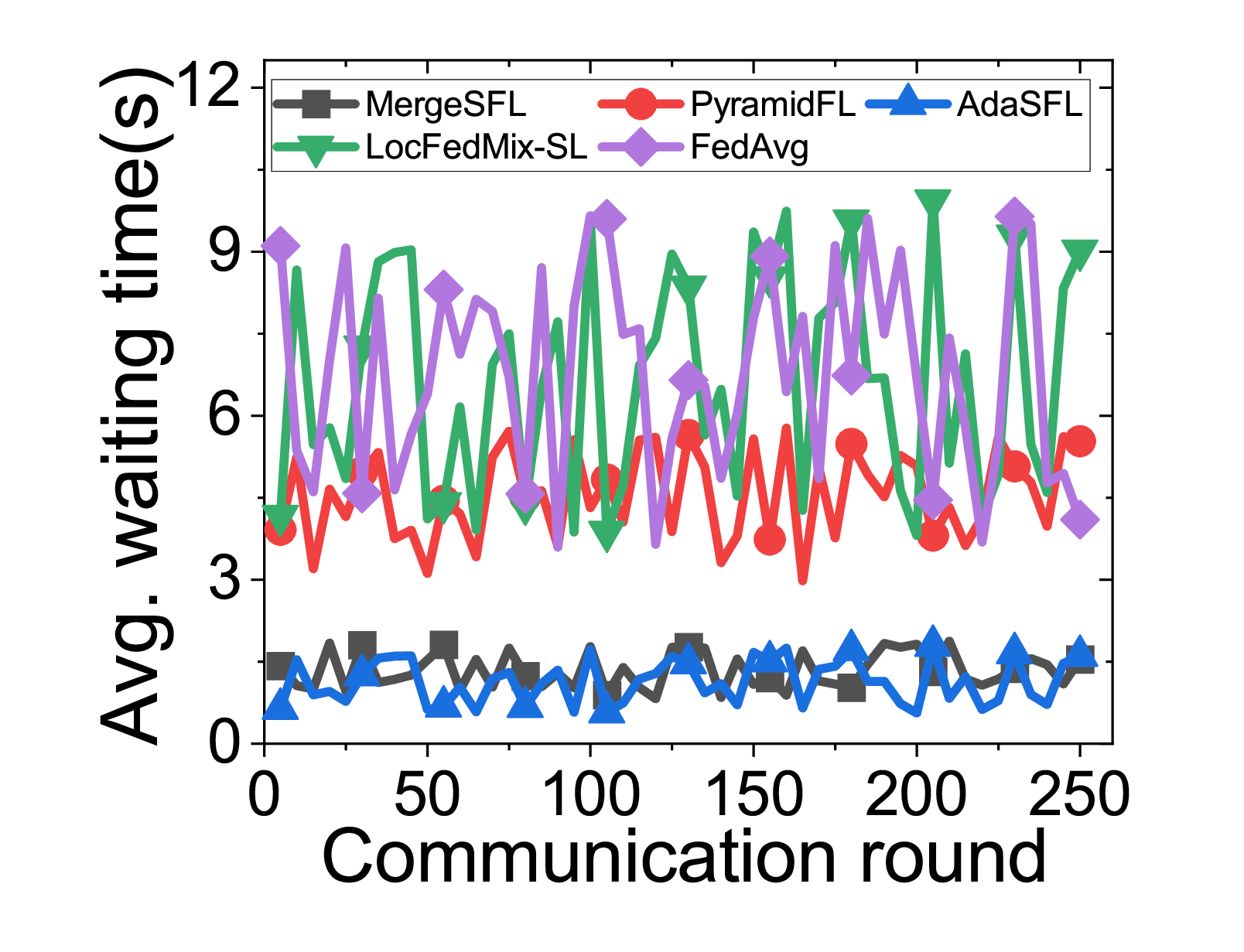}
    \label{fig:Speech-waiting_time}
}\quad 
\subfigure[CIFAR-10]
{
    \includegraphics[width=0.22\linewidth,height=3.2cm]{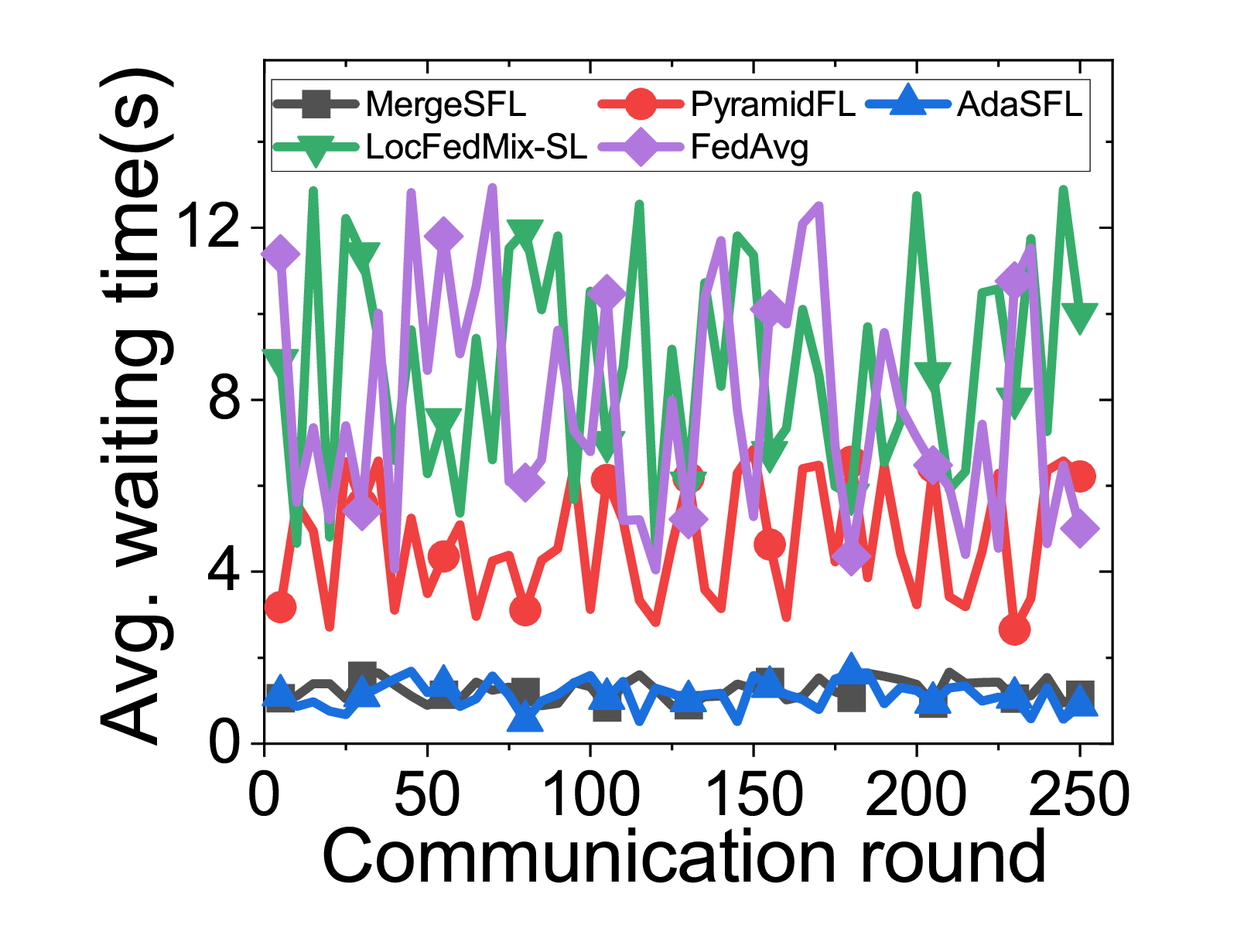}
    \label{fig:CIFAR10-waiting_time}
}\quad 
\subfigure[IMAGE-100]
{
    \includegraphics[width=0.22\linewidth,height=3.2cm]{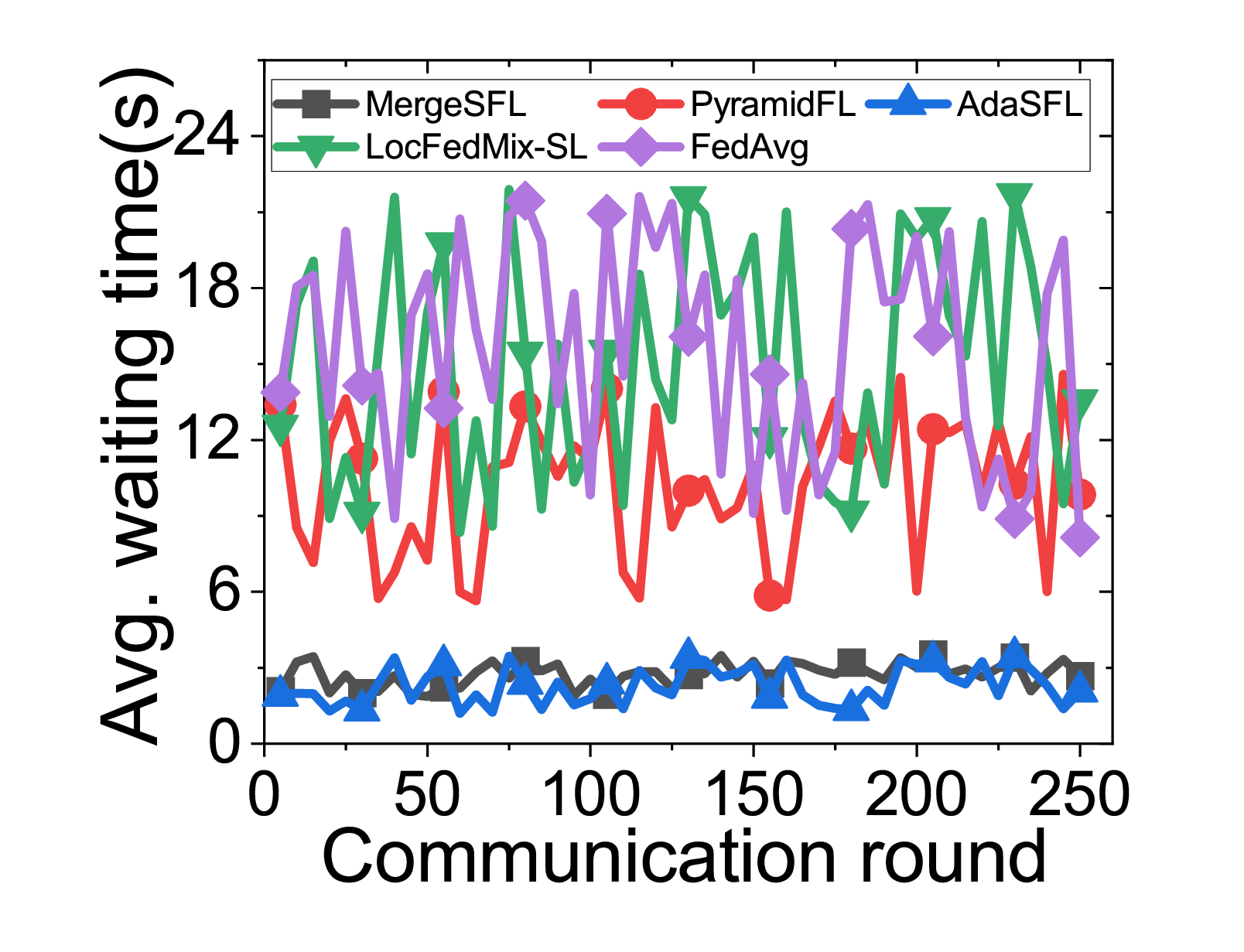}
    \label{fig:IMAGE100-waiting_time}
}
\vspace{-0.2cm}
\caption{Average waiting time of five approaches on the four datasets.}
\label{fig:waiting_time}
\vspace{-0.5cm}
\end{figure*}

4) \textbf{\textit{Image Classification}.} 
ImageNet \cite{russakovsky2015imagenet} is a dataset for image recognition that consists of 1,281,167 training images, 50,000 validation images and 100,000 test images from 1000 categories.
To adapt to the resource-constrained workers, we create a subset of ImageNet, called IMAGE-100, which contains 100 out of 1,000 categories, and each sample is resized with the shape of 64$\times$64$\times$3.
For the most complex tasks, we adopt a famous large model VGG16 with size of 321MB \cite{simonyan2014very}, which is much larger than the size of AlexNet, to classify the images in IMAGE-100.
The VGG16 consists of 13 convolutional layers with kernel of 3$\times$3, two fully-connected layers and a softmax output layer.

\textbf{Setting of Statistical Heterogeneity.}
In the experiments, training samples of each worker are drawn independently by a vector $\textbf{v}$.
To create non-IID datasets, we draw from a Dirichlet distribution \cite{hsu2019measuring, yurochkin2019bayesian}, \ie, $\textbf{v}\sim\textit{Dir}(\delta \textbf{q})$, where $\textbf{q}$ characterizes a prior class distribution, and $\delta>0$ is a concentration parameters controlling the identicalness among workers.
With $\delta \rightarrow \infty$, all workers have identical distributions to the prior class distribution (\ie, IID); with $\delta \rightarrow 0$, each worker holds data samples from only one class, which indicates high degree of statistical heterogeneity.
We specify 6 values (\eg, $\infty$, 1, 0.5, 0.25, 0.2, 0.1) for $\delta$ to generate different data distributions that cover a spectrum of identicalness, and define $p=1/{\delta}$ (\ie, $p=0,1,2,4,5,10$) to quantify the non-IID levels.
The degree of statistical heterogeneity increases as $p$ increases, and $p$ = 0 is a special case of IIDness.

\textbf{Baselines.}
We measure the effectiveness of \method through a comparison with three baselines.

1) \textbf{\textit{FedAvg}} \cite{mcmahan2017communication} is a famous FL approach that trains the entire models on all participating workers using identical batch size, and aggregates them to derive a global model.

2) \textbf{\textit{LocFedMix-SL}} \cite{oh2022locfedmix} is a typical and advanced SFL approach. 
It proposes to reduce the aggregation frequency of bottom models to save the traffic consumption, but can not fully utilize the capacities of heterogeneous workers.

3) \textbf{\textit{AdaSFL}} \cite{liao2023accelerating} is a state-of-the-art SFL approach.
It assigns adaptive and diverse batch sizes for different workers to address system heterogeneity, but still cannot deal with the statistical heterogeneity.


4) \textbf{\textit{PyramidFL}} \cite{li2022pyramidfl} is a state-of-the-art FL approach with fine-grained worker selection, and it focuses on the divergence between the selected and the unselected workers to fully exploit the computing resource and data of different workers.

\textbf{Metrics.}
We adopt the following metrics to evaluate the performance of \method and the baselines.

1) \textbf{\textit{Test Accuracy}} reflects the accuracy of the models trained by different approaches on the test datasets, and is measured by the proportion of the data correctly predicted by the models to all the test data.
Specifically, we evaluate the test accuracy of the global model (a combination of the bottom and top models in SFL) in each round, and record the final test accuracy for different approaches.

2) \textbf{\textit{Time-to-Accuracy}} is denoted as the total wall clock time taken for training a model to achieve a target accuracy (\ie, training time).
For fair comparison, we set the target accuracy as the achievable accuracy by all approaches.
We record the completion time of each round and sum up to get the total training time.
In addition, we also record the average waiting time to reflect the training efficiency of different approaches.

3) \textbf{\textit{Network Traffic}} is calculated by summing the traffic for transmitting models or features/gradients between the PS and workers when achieving a target accuracy.

\textbf{Experimental Parameters.}
By default, each set of experiments will run 150 communication rounds for CNN-H, and 250 communication rounds for CNN-S, AlexNet and VGG16.
For CNN-H, the learning rate is initialized as 0.1 and the decay rate is specified as 0.98.
The learning rates and decay rates for CNN-S, AlexNet and VGG16 are identical, and are initialized as 0.1 and 0.993 \cite{liao2023adaptive, rothchild2020fetchsgd}, respectively.
We set the local updating frequency $\tau$=10 for CNN-H, $\tau$=30 for CNN-S and AlexNet, $\tau$=40 for VGG16.
For the SFL approaches, we separately split the CNN-H, CNN-S, AlexNet, and VGG16 at the 3rd, 4th, 5th, and 13th layer.

\subsection{Overall Performance}
Firstly, we conduct a set of experiments on the IID datasets to evaluate the performance of \method and the baselines.
The training processes of these approaches are presented in Fig. \ref{fig:IID}.
By the results, all the approaches achieve the similar test accuracy eventually.
However, \method achieves the fastest convergence rate, outperforming the other approaches by a significant margin on all the four datasets.
For instance, by Fig. \ref{fig:HAR-IID}, \method takes 1,130s to achieve 87\% accuracy for CNN-H on HAR, while PyramidFL, AdaSFL, LocFedMix-SL and FedAvg consume 1,946s, 1,471s, 2,939s, 4,401s, respectively.
By Fig. \ref{fig:Speech-IID}, \method also outperforms the other approaches in terms of total completion time for CNN-S on Speech.
Besides, Fig. \ref{fig:CIFAR10-IID} shows that \method reduces the total completion time for AlexNet on CIFAR-10 by about 39\%, 24\%, 54\% and 69\%, compared to PyramidFL, AdaSFL, LocFedMix-SL and FedAvg, respectively.
Moreover, for VGG16 on IMAGE-100, as shown in Fig. \ref{fig:IMAGE100-IID}, \method can separately speed up training by about 1.74$\times$, 1.39$\times$, 2.46$\times$ and 4.14$\times$, compared to PyramidFL, AdaSFL, LocFedMix-SL and FedAvg.
These results demonstrate the superiority of \method in addressing system heterogeneity.

Secondly, we also conduct a set of experiments of these approaches on all the datasets with non-IID level $p$=10, and the results are presented in Fig. \ref{fig:non-IID}. 
We observe that \method maintains the similar convergence rate as that in the IID scenario, and achieves the highest accuracy among these approaches.
For instance, by Fig. \ref{fig:HAR-non-IID}, \method achieves 86.8\% accuracy in 1,484s for CNN-H on HAR, while PyramidFL, AdaSFL, LocFedMix-SL, and FedAvg takes 2,207s, 1,887s, 3,154s, and 4,745s to reach the accuracy of 79.44\%, 72.53\%, 72.38\%, and 72.82\%, respectively.
Similarly, Fig. \ref{fig:Speech-non-IID} illustrates that \method separately improves the final test accuracy by about 5.82\%, 25.10\%, 25.70\% and 26.22\% for CNN-S on Speech, compared to PyramidFL, AdaSFL, LocFedMix-SL and FedAvg.
Besides, by Fig. \ref{fig:CIFAR10-non-IID}, when achieving the similar test accuracy of around 60\% for AlexNet on CIFAR-10, \method reduces the total completion time by about 67\%, 73\%, 85\% and 89\%, compared to PyramidFL, AdaSFL, LocFedMix-SL and FedAvg, respectively.
Moreover, as shown in Fig. \ref{fig:IMAGE100-IID}, for VGG16 on IMAGE-100 with the same training time of 5,200s, \method improves the test accuracy by about 17.18\%, 19.68\%, 30.98\% and 45.84\%, compared to PyramidFL, AdaSFL, LocFedMix-SL and FedAvg, respectively.
These results demonstrate that \method is effective in simultaneously tackling the heterogeneity challenges with feature merging and batch size regulation.

\begin{figure*}[t]
\centering
\subfigure[HAR]
{
    \includegraphics[width=0.22\linewidth,height=3.2cm]{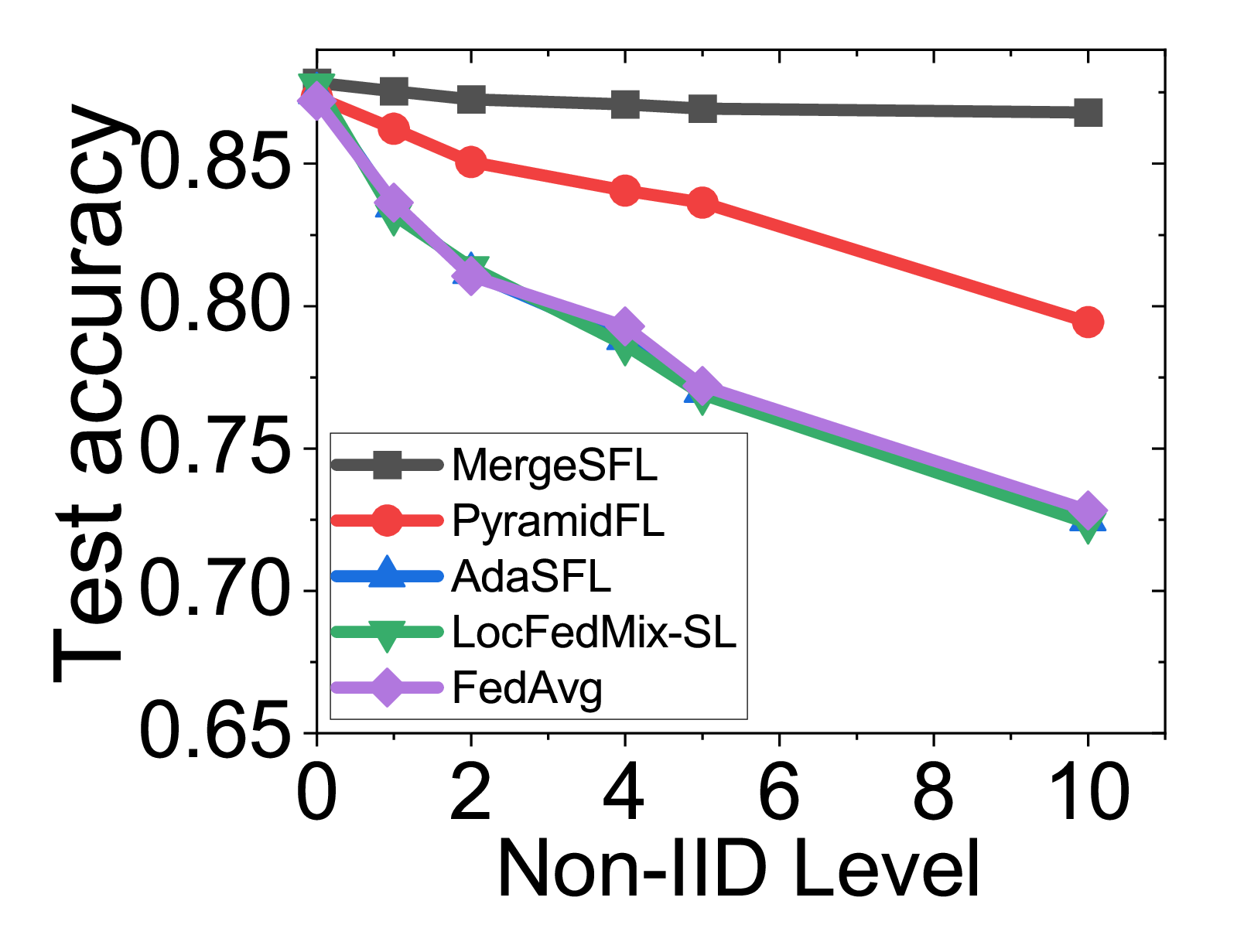}
    \label{fig:HAR-non-IID_level}
}\quad 
\subfigure[Speech]
{
    \includegraphics[width=0.22\linewidth,height=3.2cm]{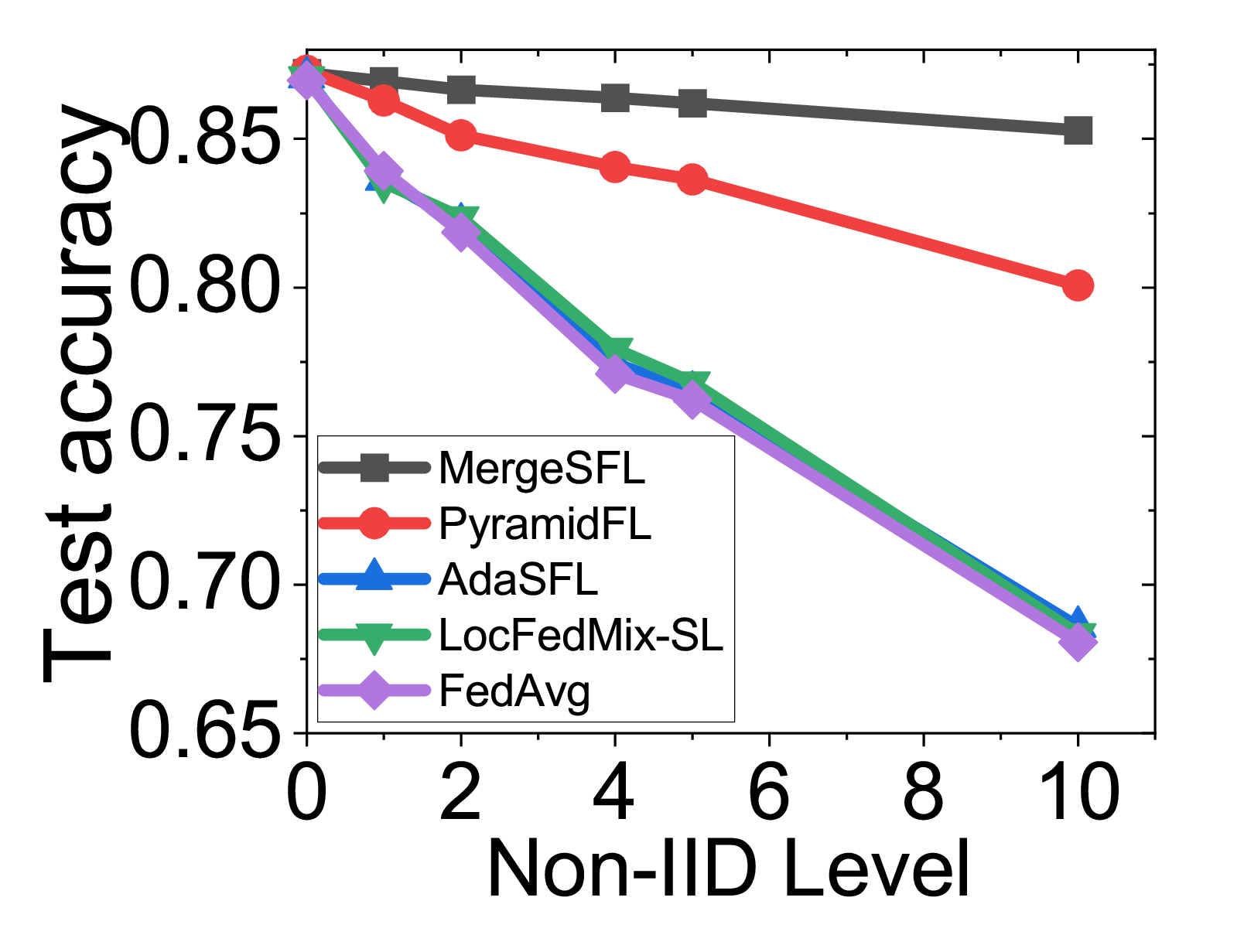}
    \label{fig:Speech-non-IID_level}
}\quad 
\subfigure[CIFAR-10]
{
    \includegraphics[width=0.22\linewidth,height=3.2cm]{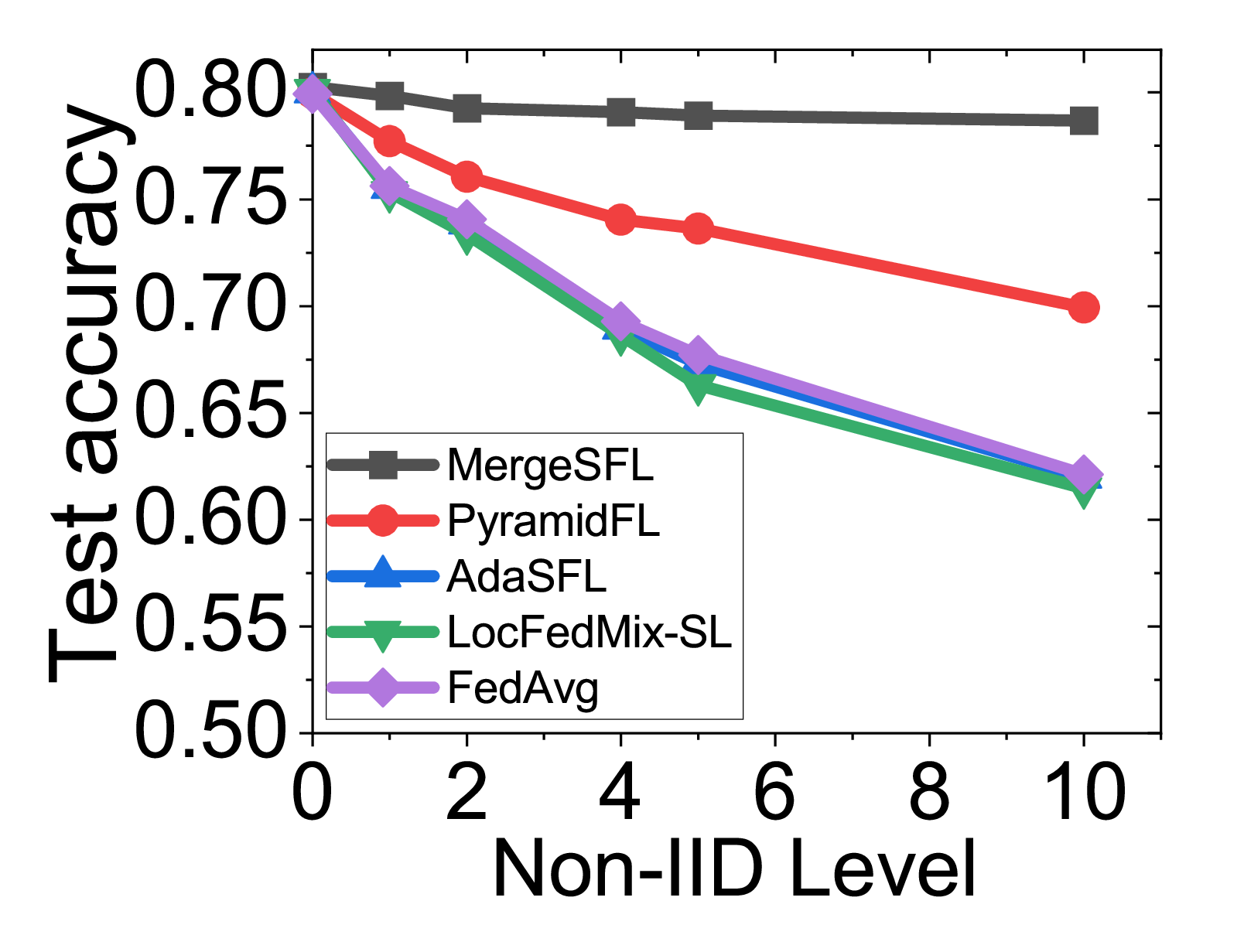}
    \label{fig:CIFAR10-non-IID_level}
}\quad 
\subfigure[IMAGE-100]
{
    \includegraphics[width=0.22\linewidth,height=3.2cm]{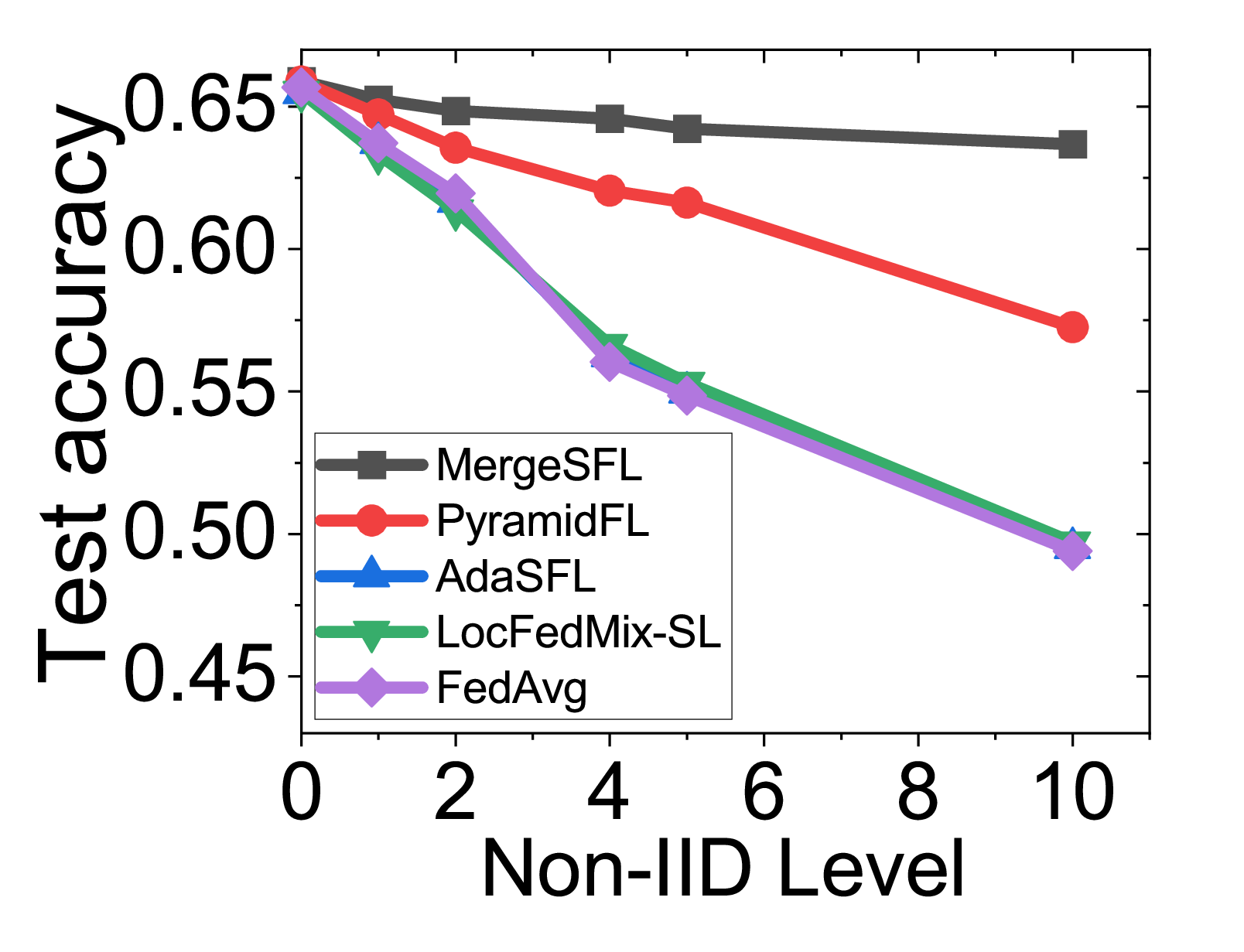}
    \label{fig:IMAGE100-non-IID_level}
}
\vspace{-0.2cm}
\caption{Test accuracy varies with different non-IID levels.}
\label{fig:non-IID_level}
\vspace{-0.4cm}
\end{figure*}

Thirdly, to illustrate the advantage of \method in saving communication resource, we present the network traffic consumption of these approaches when achieving different target accuracies in Fig. \ref{fig:bandwidth}.
By the results, the network traffic consumption of all approaches increases with the target accuracy for all the four datasets.
Furthermore, \method always consumes the least network traffic among all approaches.
In addition, model splitting (\ie, \method, AdaSFL and LocFedMix-SL) helps to save much more network traffic compared to typical FL approaches (\ie, PyramidFL and FedAvg).
AdaSFL with adaptive local updating frequency reduce the network traffic consumption while \method with adaptive worker arrangement further reduce the network traffic consumption.
As shown in Fig. \ref{fig:bandwidth_speech}, when achieving 87\% accuracy, \method, AdaSFL and LocFedMix-SL consume 1,229MB, 1,694MB and 2,398MB, respectively, while PyramidFL and FedAvg consume 3,397MB and 4,036MB for CNN-S on Speech.
Besides, as illustrated in Fig. \ref{fig:bandwidth_image}, \method saves network traffic consumption by about 49\%, 19\%, 38\% and 58\% when achieving 65\% accuracy for VGG16 on IMAGE-100, compared to the baselines (\ie, PyramidFL, AdaSFL, LocFedMix-SL and FedAvg).

\begin{figure}[!t]
\centering
\subfigure[IID]
{
    \includegraphics[width=0.45\linewidth,height=3.2cm]{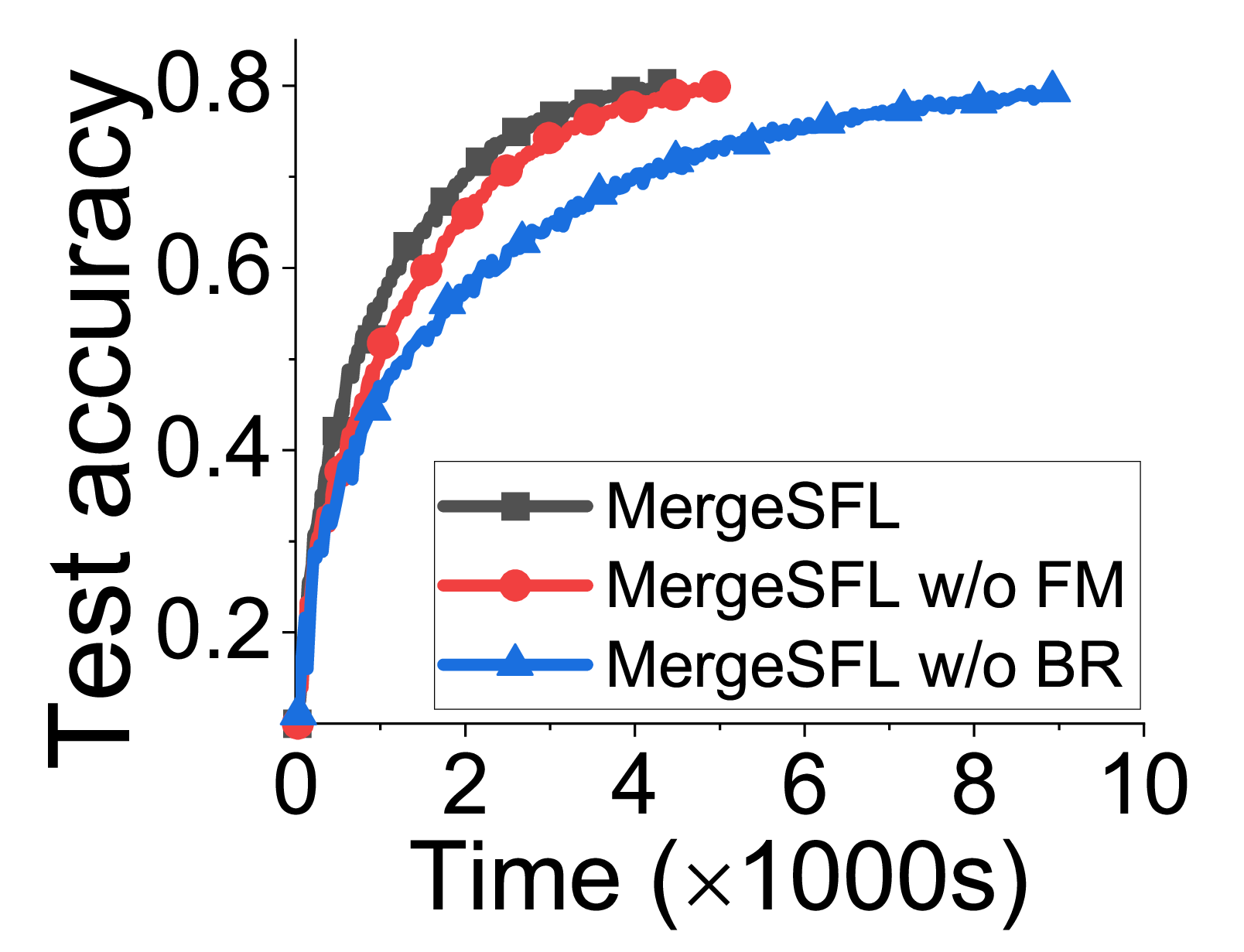}
    \label{fig:key_IID}
}\quad
\subfigure[Non-IID]
{
    \includegraphics[width=0.45\linewidth,height=3.2cm]{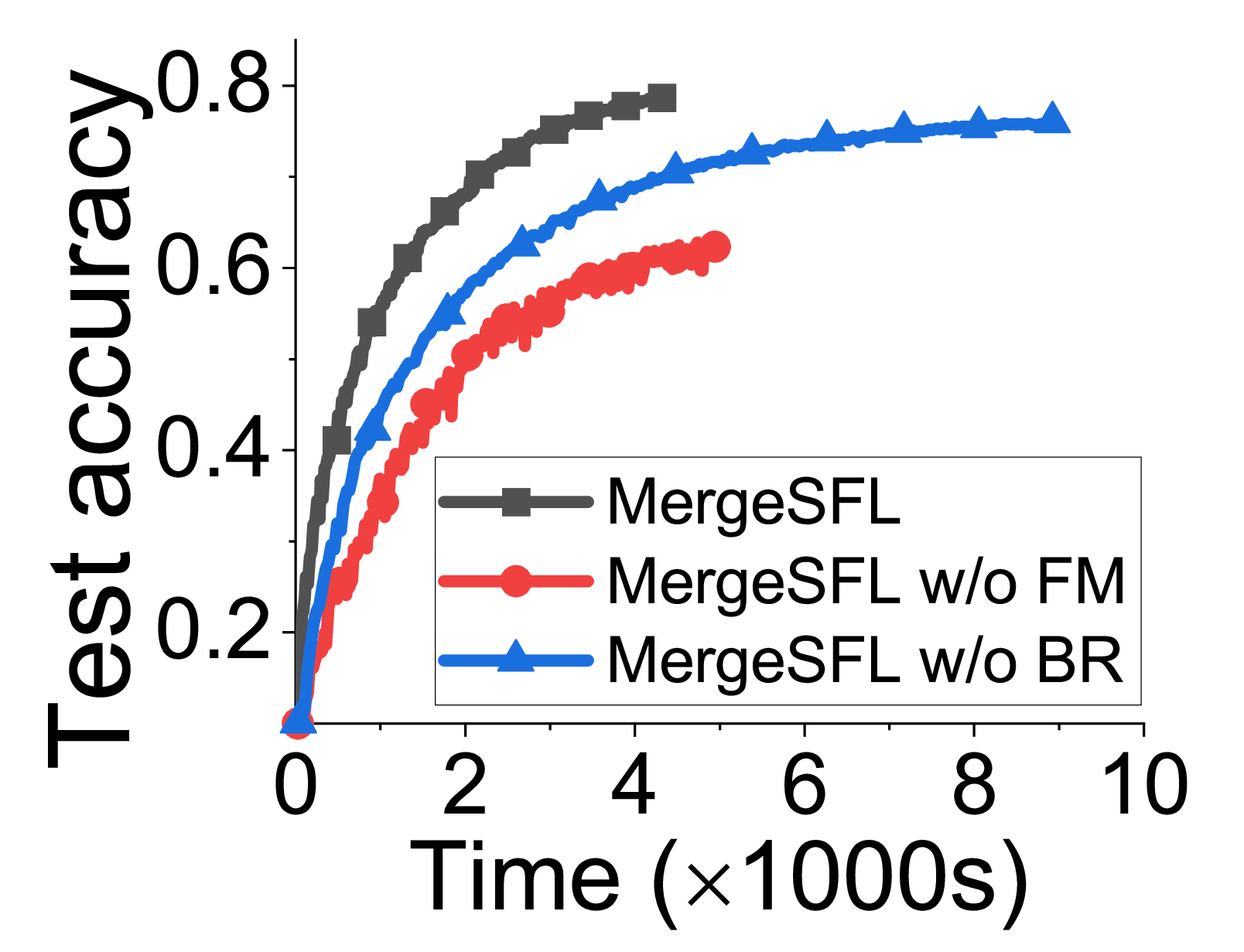}
    \label{fig:key_nonIID}
}
\vspace{-0.4cm}
\caption{Effects of feature merging and batch size regulation.}
\label{fig:key_strategies}
\vspace{-0.6cm}
\end{figure}

To further demonstrate the robustness of \method towards system heterogeneity, we illustrate the average waiting time of the five approaches on the four datasets in Fig. \ref{fig:waiting_time}.
AdaSFL with adaptive and diverse batch sizes for heterogeneous workers achieves the least waiting time, but the waiting time of \method is close to AdaSFL and is much less than that of other approaches.
For instance, by Fig. \ref{fig:HAR-waiting_time}, the average waiting time of \method is 1.2s for CNN-H on HAR while PyramidFL, AdaSFL, LocFedMix-SL and FedAvg incur average waiting time of 3.4s, 1.1s, 5.9s and 6.1s, respectively.
Considering the workers with varying capacities, LocFedMix-SL and FedAvg use fixed and identical batch size for model training, without considering system heterogeneity, thus lead to non-negligible waiting time.
PyramidFL with adaptive worker selection to fully exploit the computing resource and data of different workers reduces the average waiting time to a certain extent.
Concretely, by Fig. \ref{fig:IMAGE100-waiting_time}, \method can reduce the average waiting time to train VGG16 on IMAGE-100 by about 65\%, 79\% and 81\%, compared to PyramidFL, LocFedMix-SL and FedAvg.

\begin{figure}[!t]
\centering
\subfigure[Completion Time]
{
    \includegraphics[width=0.45\linewidth,height=3.2cm]{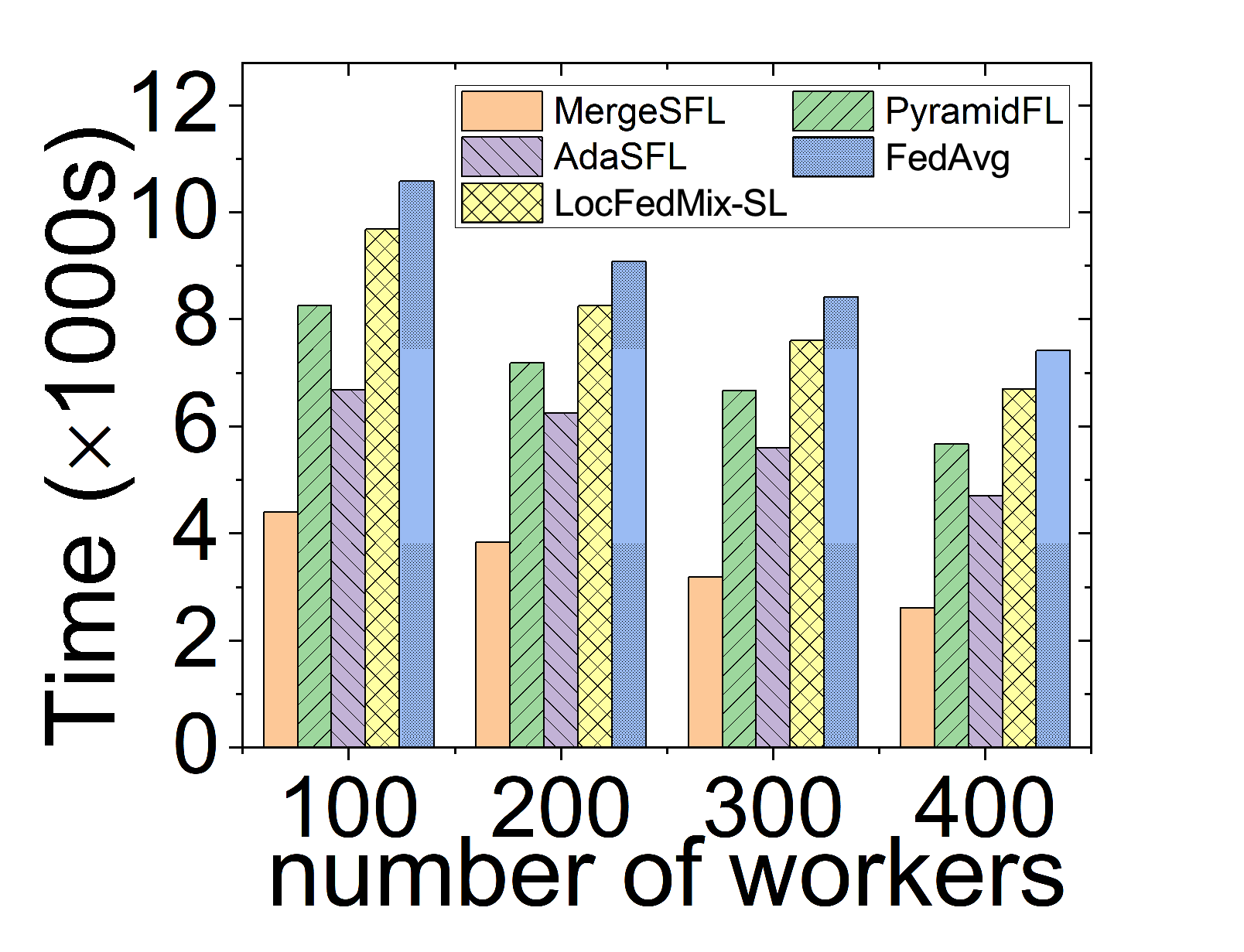}
    \label{fig:scales_time}
}\quad
\subfigure[Training Process]
{
    \includegraphics[width=0.45\linewidth,height=3.2cm]{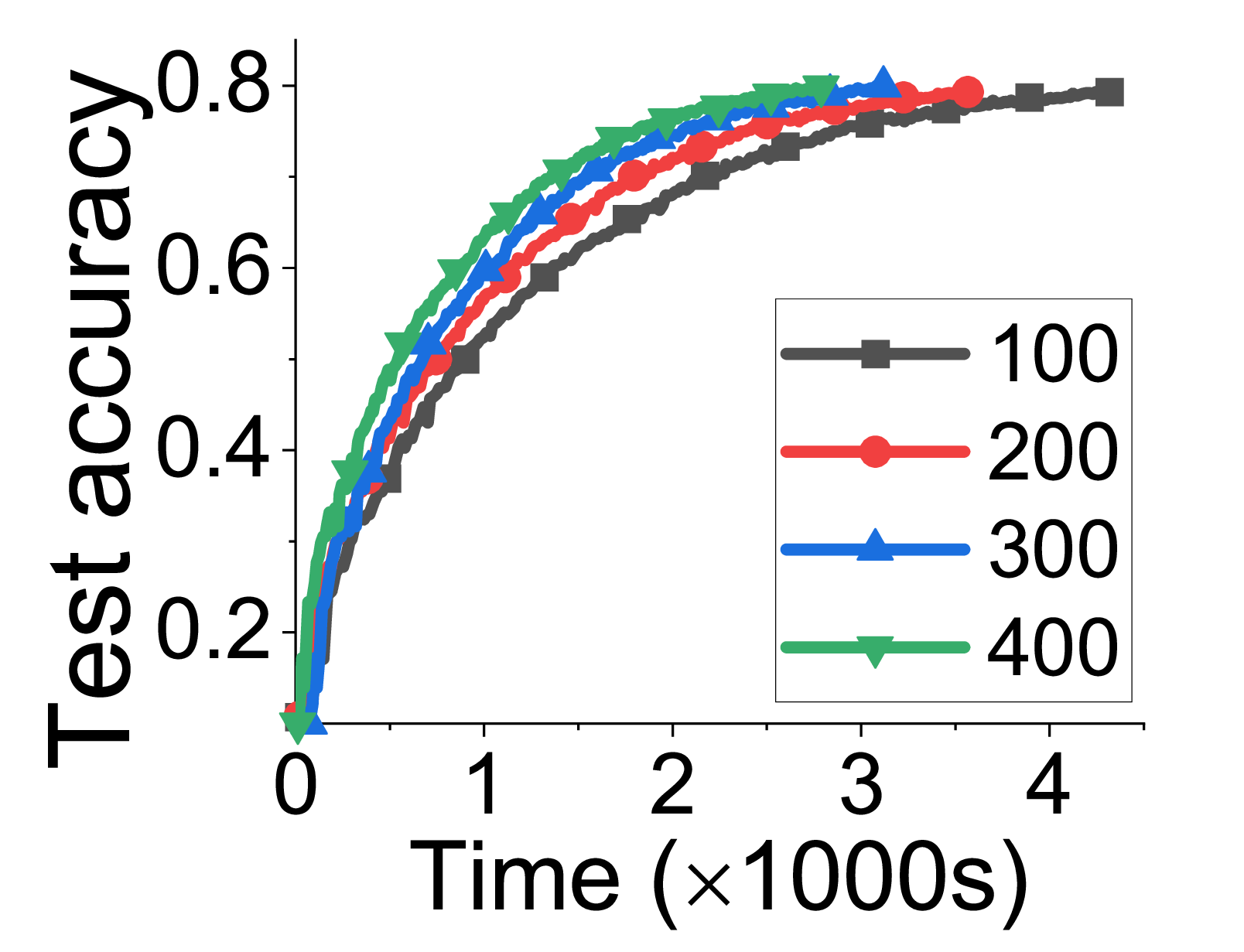}
    \label{fig:scales_time_acc}
}
\vspace{-0.4cm}
\caption{Performance comparison with different number of workers.}
\label{fig:scales}
\vspace{-0.6cm}
\end{figure}

\subsection{Effect of Non-IID Levels}
To demonstrate the effectiveness of \method in handling non-IID data, we present the test accuracy of different approaches at varying non-IID levels in Fig. \ref{fig:non-IID_level}, where the horizontal axis denotes the non-IID level of the datasets.
As shown in Fig. \ref{fig:non-IID_level}, the test accuracy of the models trained by the five approaches on all datasets decreases as the non-IID level increases.
However, \method consistently outperforms the other approaches on all datasets.
LocFedMix-SL, AdaSFL and FedAvg, without considering the challenges of system and statistical heterogeneity, exhibit the lowest model accuracy on non-IID datasets.
PyramidFL, which focuses on the divergence between the selected workers and the remaining to fully exploit the computing resource and data of different workers, can mitigate the impact of non-IID data on model training to some extent.
Specifically, by Fig. \ref{fig:HAR-non-IID_level}, with non-IID level of $p$=10 on HAR, \method and PyramidFL achieve 86.8\% and 79.44\% accuracy, while LocFedMix-SL, AdaSFL and FedAvg only achieve 72.53\%, 72.38\% and 72.82\% accuracy.
Moreover, as shown in Fig. \ref{fig:Speech-non-IID_level}, while transitioning from IID to non-IID level of 
$p$=10 on Google Speech, \method and PyramidFL suffer from only 1.93\% and 7.23\% loss in accuracy, while the accuracy loss for AdaSFL, LocFedMix-SL and FedAvg is 18.54\%, 18.74\% and 18.91\%, respectively.
Notably, by Fig. \ref{fig:CIFAR10-non-IID_level}, \method can achieve improvement of test accuracy by about 12.50\%, 27.36\%, 27.98\%, 26.66\% on CIFAR-10 with non-IID level of $p$=10, compared to the baselines (\ie, PyramidFL, AdaSFL, LocFedMix-SL, FedAvg).

\subsection{Effect of Key Strategies}
There are two key strategies, \ie, feature merging and batch size regulation, that are developed to enhance the performance of SFL.
Herein, we conduct several sets of experiments for AlexNet on CIFAR-10 with IID distribution ($p$=0) and non-IID distribution ($p$=10) to evaluate the effectiveness of the two critical strategies.
We adopt the \method without feature merging (\method w/o FM) and \method without batch size regulation (\method w/o BR) as the baselines.
Concretely, in \method w/o FM, the PS directly applies the features of workers with diverse batch sizes to separately perform forward/backward propagation without feature merging.
While in \method w/o BR, all workers are assigned with an identical batch size, that is the average of batch sizes in \method, for feature merging and model training.
By Fig. \ref{fig:key_strategies}, \method w/o FM converges as fast as \method on the IID dataset, while \method w/o BR achieves similar test accuracy as \method on the non-IID dataset.
Powered by feature merging and batch size regulation, \method can speed up training by about 2.17$\times$ compared to \method w/o BR, and improve the final test accuracy by about 28.83\% compared to \method w/o FM, which reflects the positive roles of the two strategies.

\vspace{-0.1cm}
\subsection{Effect of System Scales}
In this section, to demonstrate the robustness of \method, we evaluate the performance of \method and baselines with different scales of participating workers.
We train AlexNet on CIFAR-10 with four scales (\ie, 100, 200, 300, 400) through extensive simulation experiments, which are conducted on an AMAX deep learning workstation equipped with an Intel(R) Xeon(R) Gold 5218R CPU, 8 NVIDIA GeForce RTX 3090 GPUs and 256 GB RAM.
The results of completion time to achieve 80\% accuracy for these approaches are presented in Fig. \ref{fig:scales_time}, while the training processes of different scales for \method are presented in Fig. \ref{fig:scales_time_acc}.
As the number of participating workers increases, all approaches achieve faster convergence.
For instance, \method with 400 workers achieves a speedup of 1.68$\times$, 1.47$\times$ and 1.23$\times$, compared to \method with 100, 200 and 300 workers, respectively.
The reason is that the number of samples on a worker is limited and more workers contribute more local data for training.
In addition, \method reaches the target accuracy 1.47$\times$$\sim$2.85$\times$ faster than the baselines (\ie, PyramidFL, AdaSFL, LocFedMix-SL, FedAvg) regarding the different scales of workers.


\vspace{-0.1cm}
\section{Related Work}\label{sec:relwork}
The existing split federated learning (SFL) researches are initially proposed to offload the computing tasks on resource-constrained workers when training large-scale DL models, but they are unable to simultaneously overcome the system and statistical heterogeneity \cite{thapa2022splitfed, han2021accelerating, pal2021server,liao2023accelerating,oh2022locfedmix}.
For instance, Thapa \etal \cite{thapa2022splitfed} demonstrate the feasibility of SFL and pioneer the first SFL method, termed SplitFed, which aggregates bottom models after each local updating.
Such frequent aggregation results in high network traffic consumption.
To save the traffic consumption, Han \etal \cite{han2021accelerating} propose LocSplitFed and allow the workers not send features to the PS by using local-loss-based training. 
Then, Oh \etal \cite{oh2022locfedmix} propose LocFedMix-SL, which is implemented to maintain all the benefits of SplitFed and LocSplitFed with multiple local updating frequency, but cannot fully utilize the capacities of heterogeneous workers.
Although Liao \etal \cite{liao2023accelerating} propose an advanced solution AdaSFL to assign adaptive and diverse batch sizes for different workers, AdaSFL still cannot address the statistical heterogeneity.
Despite these notable advancements, none of the existing SFL works have yet explored to simultaneously tackle heterogeneity issues.

Prior to the emergence of SFL, the system and statistical heterogeneity issues have been studied and addressed in many typical FL works \cite{li2022data,xie2023federatedscope, arisdakessian2023towards, lai2021oort, shin2022fedbalancer}.
On one hand, in order to alleviate the negative effect of system heterogeneity, some works \cite{wang2019adaptive, xu2022adaptive, tyagi2020taming, ma2021adaptive, liu2019accelerate} investigate to optimize the local updating frequencies and batch sizes of different workers.
For example, Xu \etal \cite{xu2022adaptive} propose FedLamp to assign the relatively high-performance workers (with high computing/communication capacities) with larger local updating frequencies.
Besides, Ma \etal \cite{ma2021adaptive} propose to assign adaptive batch sizes and scaled learning rates for heterogeneous workers.
On the other hand, some works \cite{shin2022fedbalancer, li2021sample, tuor2021overcoming} actively selects high-utility data samples to address statistical heterogeneity.
For instance, Li \etal \cite{li2021sample} propose to prioritize client training samples with higher importance in FL, while Shin \etal \cite{shin2022fedbalancer} propose FedBalancer, which introduce a deadline control strategy to optimize the time-to-accuracy performance.
In addition, other works \cite{luo2022tackling,li2022pyramidfl,lai2021oort} propose to employ worker selection to simultaneously address system and statistical heterogeneity. 
Specifically, Li \etal \cite{li2022pyramidfl} develop PyramidFL, a fine-grained worker selection strategy that focuses on the divergence between the selected workers and the remaining workers to fully exploit the computing resource and data of different workers.
However, those FL researches may be infeasible if the resource-constrained workers do not have enough memory to run the program of training large-scale models.
Moreover, the relevant optimization techniques are probably unable to be directly applied for SFL and \method, since workers in SFL maintain only the bottom models and must continuously exchange features/gradients with the PS that possesses the top model.

\vspace{-0.1cm}
\section{Conclusion}\label{sec:conclusion}

In this paper, we have designed and implemented a novel SFL framework, termed \method, which incorporated feature merging and batch size regulation to address the system and statistical heterogeneity.
By assigning diverse as well as suitable batch sizes for heterogeneous workers, and merging the features from workers into the mixed feature sequence, \method could promote model accuracy and training efficiency for SFL.
The experimental results showed that \method significantly outperformed the baselines, providing a speedup of 1.39$\times$$\sim$4.14$\times$ with the improvement of final model accuracy by 5.82\%$\sim$26.22\%.

\vspace{-0.1cm}
\section*{Acknowledgment}
This article is supported in part by the National Key Research and Development Program of China (Grant No. 2021YFB3301500); in part by the National Science Foundation of China (NSFC) under Grants 61936015, 62102391 and 62132019; in part by the Jiangsu Province Science Foundation for Youths (Grant No. BK20210122).


\balance
\bibliographystyle{IEEEtran}
\bibliography{MergeSFL.bib}

\end{document}